\documentclass[pmlr,twocolumn,10pt]{jmlr} % W&CP article

\usepackage{booktabs}
\usepackage{multirow}
\usepackage{multicol}
\usepackage{algorithm}
\usepackage{algorithmic}
\usepackage[capitalize,noabbrev]{cleveref}

\usepackage[switch]{lineno}

\newcommand{\cs}[1]{\texttt{\char`\\#1}}% remove this in your real article

\theorembodyfont{\upshape}
\theoremheaderfont{\scshape}
\theorempostheader{:}
\theoremsep{\newline}
\newtheorem*{note}{Note}

\jmlrvolume{333}
\jmlryear{2026}
\jmlrsubmitted{LEAVE UNSET}
\jmlrpublished{LEAVE UNSET}
\jmlrworkshop{Conference on Health, Inference, and Learning (CHIL) 2026} % W&CP title

\title[LapDDPM: Spectral Perturbation Diffusion for Robust Single-Cell Manifold Generation]{LapDDPM: Spectral Perturbation Diffusion for Robust Single-Cell Manifold Generation}

\author{%
  \Name{Lorenzo Bini} \Email{lorenzo.bini@unige.ch} %\and 
  \\
  \Name{Stéphane Marchand-Maillet} \Email{stephane.marchand-maillet@unige.ch}\\
  \addr Department of Computer Science, University of Geneva, Geneva, Switzerland.
 }

\begin{document}
\maketitle

\begin{abstract}
Generating high-fidelity and biologically plausible synthetic single-cell RNA sequencing (scRNA-seq) data is a critical challenge in computational biology, driven by the need to model high-dimensional, sparse, and non-linear cellular manifolds. Existing generative models often fail to capture the complex topology of cellular differentiation or lack robustness against technical noise and structural variability. We introduce LapDDPM, a novel conditional Graph Diffusion Probabilistic Model designed for robust manifold learning and high-fidelity generation. LapDDPM integrates graph-based inductive biases with score-based generative modeling, enhanced by a novel spectral adversarial perturbation mechanism. By systematically perturbing graph edge weights along principal spectral modes during training, our method acts as a Distributionally Robust Optimization (DRO) framework, enforcing invariance to structural noise. We further extend LapDDPM to spatial transcriptomics and multi-modal data, treating generation as a robust inverse problem on cellular graphs. Extensive experiments on diverse datasets, including PBMC3K, Dentate Gyrus, HLCA, Visium, and 10x Multiome, demonstrate that LapDDPM significantly outperforms state-of-the-art baselines in distribution matching, manifold preservation, and downstream utility, generating biologically coherent cell states.
\end{abstract}

\paragraph*{Data and Code Availability}
This paper uses publicly available scRNA-seq datasets including PBMC3K, Dentate Gyrus, Tabula Muris, and the Human Lung Cell Atlas (HLCA), with corresponding citations provided in the main text. Code is publicly available at \url{https://github.com/LorenzoBini4/laplace-DDPM}.

\paragraph*{Institutional Review Board (IRB)}
This study uses only publicly available, de-identified datasets and does not involve human subjects research; therefore, IRB approval was not required.

\section{Introduction}
\label{introduction}
Single-cell RNA sequencing (scRNA-seq) has revolutionized our understanding of cellular heterogeneity, enabling the characterization of gene expression at an unprecedented resolution \citep{tang2009rna, saliba2014single}. However, the generation of scRNA-seq data is often resource-intensive, limited by sample availability, and inherently noisy due to biological and technical variations \citep{kiselev2019challenges}. This has spurred significant interest in developing computational methods for generating synthetic scRNA-seq data that closely mimics real biological samples \citep{eraslan2019single}. Such synthetic data can serve various purposes, including data augmentation for rare cell types, benchmarking of computational tools, and in-silico perturbation studies.

Recent advancements in generative modeling, particularly Diffusion Probabilistic Models (DDPMs) \citep{ho2020denoising}, have shown remarkable success in synthesizing high-fidelity data across various domains. Concurrently, flow-based generative models \citep{dinh2014nice, dinh2016density, cfGen} have also gained prominence for their ability to learn complex data distributions through invertible transformations, offering exact likelihood estimation. The application of Graph Neural Networks (GNNs) has also gained traction in single-cell biology, leveraging the inherent graph-like structures of cellular relationships (e.g., cell-cell similarity networks or lineage trees) \citep{zhao2021graph, zhou2022graph}. Integrating these powerful paradigms presents a promising avenue for generating scRNA-seq data, where GNNs can capture complex cellular interactions and DPMs or flow models can model the intricate gene expression distributions.

Despite these advances, existing generative models for scRNA-seq data often face several limitations. Many struggle with the unique characteristics of scRNA-seq data, such as its high dimensionality, sparsity (due to dropout events), zero-inflation, and complex, non-Gaussian distributions \citep{kiselev2019challenges}. Furthermore, while some methods incorporate graph information, they may lack mechanisms to ensure robustness against structural noise or subtle variations in cell-cell interaction networks, which are common in real biological contexts. The ability to generate data conditioned on specific biological metadata (e.g., cell type, tissue origin) is also crucial for practical applications, yet remains a challenge for many unconditional models.

To address these limitations, we propose a novel conditional generative framework for scRNA-seq data. LapDDPM (Laplacian Diffusion Denoising Probabilistic Model) integrates graph-based representations with a score-based diffusion model, further enhanced by a unique spectral adversarial perturbation mechanism applied directly to the graph's edge weights. Our method aims to synthesize realistic scRNA-seq count matrices that faithfully reproduce the statistical properties and cellular heterogeneity of real data, conditionally on cell types.
Our approach makes three key contributions:
\begin{itemize}
    \item We propose a \textbf{robust graph-based representation learning framework} for scRNA-seq that leverages k-Nearest Neighbors (k-NN) graphs and Laplacian Positional Encodings (LPEs) to inject structural inductive biases into the generative process.
    \item We develop a \textbf{conditional score-based diffusion model} operating in a learned latent space, capable of modeling the complex, high-dimensional manifolds of cellular states while conditioning on biological metadata.
    \item We introduce a novel \textbf{spectral adversarial perturbation mechanism} as a principled Distributionally Robust Optimization (DRO) objective. By perturbing edge weights along the principal spectral modes of the graph, we enforce invariance to structural noise, enhancing generalization across batch effects and biological variability.
\end{itemize}
Extensive experiments across diverse scRNA-seq datasets, including PBMC3K, Dentate Gyrus, Tabula Muris, and the Human Lung Cell Atlas (HLCA), comprehensively validate LapDDPM's effectiveness. Beyond superior quantitative metrics (RBF-kernel MMD and 2-Wasserstein distance), LapDDPM preserves marker-gene structure and pseudotime ordering while improving downstream utility through stronger TSTR and GRN fidelity, underscoring its value for conditionally synthesizing biologically plausible, cell-type-specific scRNA-seq profiles.

\section{Related Work}
\label{sec:related_work}
Our work bridges generative modeling for scRNA-seq, Graph Neural Networks (GNNs), and diffusion models, specifically addressing robust manifold learning.

\subsection{Generative Models for scRNA-seq Data}
Early approaches adapted general machine learning models like Variational Autoencoders (VAEs) \citep{lopez2018deep, eraslan2019single} and Generative Adversarial Networks (GANs) \citep{ghosh2020generating, li2021adversarial} to scRNA-seq. VAEs learn low-dimensional latent representations, often accounting for sparsity, while GANs map simple priors to data distributions via adversarial training. More recently, flow-based models \citep{chen2021flow} offered exact likelihood estimation. However, these methods often struggle to capture intricate topological heterogeneities or ensure robustness against noise. Our work extends these efforts by leveraging graph structures and scoring-based diffusion for fidelity and robustness.

\subsection{Graph Neural Networks in Single-Cell Biology}
The inherent relational nature of single-cell data, where cells can be viewed as nodes in a graph connected by biological similarity (e.g., gene expression similarity, spatial proximity, or lineage relationships), makes GNNs a natural fit for analysis. GNNs have been applied to various tasks in single-cell biology, including cell type annotation \citep{wang2021graph}, trajectory inference \citep{qi2021graph}, and spatial transcriptomics analysis \citep{zhou2022graph}. These applications typically use GNNs as feature extractors or classifiers. Our work utilizes GNNs within a generative framework, specifically a spectral encoder, to effectively process the graph-structured representation of scRNA-seq data and extract meaningful latent features for the diffusion process.

\subsection{Diffusion Probabilistic Models}
Diffusion Probabilistic Models (DPMs) \citep{sohl-dickstein2015deep, ho2020denoising, song2020score} have emerged as a powerful class of generative models, demonstrating state-of-the-art performance across various domains, including image synthesis \citep{dhariwal2021diffusion} and audio generation \citep{kong2021diffwave}. DPMs define a forward diffusion process that gradually adds noise to data and a reverse process that learns to denoise it. The reverse process is typically modeled by a neural network that estimates the score function (gradient of the log-probability density). Recent advancements, such as score-based generative models \citep{song2020score}, have unified various DPM formulations. In the biological domain, DPMs have shown promise for protein design \citep{anand2022protein} and drug discovery \citep{hoogeboom2022equivariant}. LapDDPM builds upon the success of score-based diffusion models by adapting them to the unique characteristics of scRNA-seq data within a graph-aware latent space, enabling high-fidelity conditional generation.

\subsection{Adversarial Training and Robustness on Graphs}
Adversarial training has been widely adopted to enhance the robustness of machine learning models against malicious perturbations \citep{goodfellow2014explaining}. In the context of GNNs, adversarial attacks can target node features or the graph structure itself (e.g., adding or removing edges) \citep{zugner2018adversarial, dai2018adversarial}. Correspondingly, defense mechanisms include robust aggregation functions \citep{li2019robust} and adversarial training techniques \citep{kong2020flag,feng2020grand,thorpe2022grand++}. Recent works have explored more efficient adversarial training schemes, such as virtual adversarial training \citep{virtual-zhuo2023propagation} and diffusion-based approaches \citep{gosch2024adversarial}. Our work introduces a novel spectral adversarial perturbation mechanism that directly modifies the edge weights of the input graph during training. This approach is distinct from traditional feature or hidden representation perturbations and aims to enhance the encoder's robustness to structural variations, which is particularly relevant for the complex and often noisy graph structures derived from biological data. In Section \ref{sec:experiment}, we assess the robustness of our methods against adversarially poisoned input graphs, focusing on representations learned from graphs compromised by various structural attack strategies, including random attacks, DICE \citep{dice-waniek2018hiding}, GF-Attack \citep{gf-attack_chang2020restricted}, and Mettack \citep{mettack_gosch2024adversarial}.

\subsection{Spectral Graph Methods}
Spectral graph theory provides a powerful framework for analyzing graph properties through the eigenvalues and eigenvectors of graph matrices, such as the adjacency matrix or Laplacian \citep{chung1997spectral}. In GNNs, spectral methods have informed the design of convolutional layers \citep{defferrard2016convolutional} and provided theoretical insights into message passing \citep{balcilar2021analyzing}. More recently, spectral properties have been leveraged for graph augmentation \citep{spectral-aug-gcl-ghose2023-AAAI} and for improving the robustness of GNNs \citep{sp2gcl_bo2024graph}. LapDDPM extensively utilizes spectral graph theory by incorporating Laplacian Positional Encodings (LPEs) as node features, which provide crucial structural context to the GNN encoder. Furthermore, our novel spectral adversarial perturbation mechanism is fundamentally rooted in spectral properties, specifically leveraging the principal eigenvector of the graph's adjacency matrix to generate meaningful structural variations. This allows our model to learn robust representations by challenging it with perturbations that target the graph's dominant spectral modes.

Our work integrates these distinct yet complementary research areas, proposing LapDDPM as a unified framework for robust and conditional scRNA-seq data generation. By combining graph-based representations, spectral positional encodings, a score-based diffusion model, and spectral adversarial perturbations, LapDDPM offers a powerful solution for synthesizing high-fidelity biological data.

\section{Methodology}
\label{sec:methodology}
We detail the components of LapDDPM, a novel conditional generative framework for single-cell RNA sequencing (scRNA-seq) data. The framework leverages graph-based representations, spectral positional encodings, and a score-based diffusion model to synthesize realistic scRNA-seq count matrices conditioned on cell types. A key innovation is the integration of spectral adversarial perturbations applied directly to the graph structure during training, enhancing the model's robustness to structural variations inherent in biological data.

LapDDPM comprises three primary components: (1) a graph-based data representation module that constructs a k-NN graph from scRNA-seq data and computes LPEs, (2) a spectral encoder-decoder pair that operates on graph-structured data, and (3) a conditional score-based diffusion model that learns to generate latent representations. The overall training procedure combines diffusion, reconstruction, and KL divergence losses, with the encoder being trained on graphs perturbed by a novel spectral adversarial mechanism.

\subsection{Graph-based scRNA-seq Data Representation}
\label{sec:data_representation}
Given a scRNA-seq dataset consisting of $N$ cells and $D$ genes, represented as a count matrix $\mathbf{X} \in \mathbb{R}^{N \times D}$, we first preprocess the data to construct a graph $\mathcal{G} = (V, E)$ where nodes $V$ correspond to individual cells and edges $E$ represent cellular proximity.

\paragraph{Gene Filtering and Normalization:} Prior to graph construction, genes expressed in fewer than a specified threshold of cells are filtered out to reduce sparsity and computational burden. The raw count data is then normalized and log-transformed for stable numerical operations during feature extraction.

\paragraph{Laplacian Positional Encoding (LPE):} Firstly, a k-NN graph is constructed on the cells. To capture biologically meaningful relationships and reduce the dimensionality of the feature space for graph construction, Principal Component Analysis (PCA) is applied to the log-transformed gene expression data. The k-NN graph is then built using Euclidean distance in this PCA-reduced space, preserving local micro-environments, including rare populations, while reducing hub-node bias in noisy count space. For each cell (node), its $k$ nearest neighbors are identified, and edges are formed between them. The resulting adjacency matrix is denoted as $\mathbf{A} \in \{0,1\}^{N \times N}$.

Then, to provide the GNN backbone with positional information and enhance its ability to distinguish nodes based on their structural roles, we compute LPEs. The LPEs are derived from the eigenvectors of the normalized graph Laplacian. The normalized Laplacian $\mathbf{L}_{\text{norm}}$ is defined as:
$$\mathbf{L}_{\text{norm}} = \mathbf{I} - \mathbf{D}^{-1/2}\mathbf{A}\mathbf{D}^{-1/2}$$
where $\mathbf{I}$ is the identity matrix and $\mathbf{D}$ is the diagonal degree matrix of $\mathbf{A}$. We compute the first $k$ non-trivial eigenvectors of $\mathbf{L}_{\text{norm}}$ (excluding the trivial eigenvector corresponding to eigenvalue 0 for connected graphs), which form the LPE matrix $\mathbf{P} \in \mathbb{R}^{N \times k}$. These LPEs are then concatenated with the gene expression features for the encoder.

\subsection{LapDDPM Architecture}
\begin{figure*}[t]
    \centering
    \includegraphics[width=0.8\textwidth]{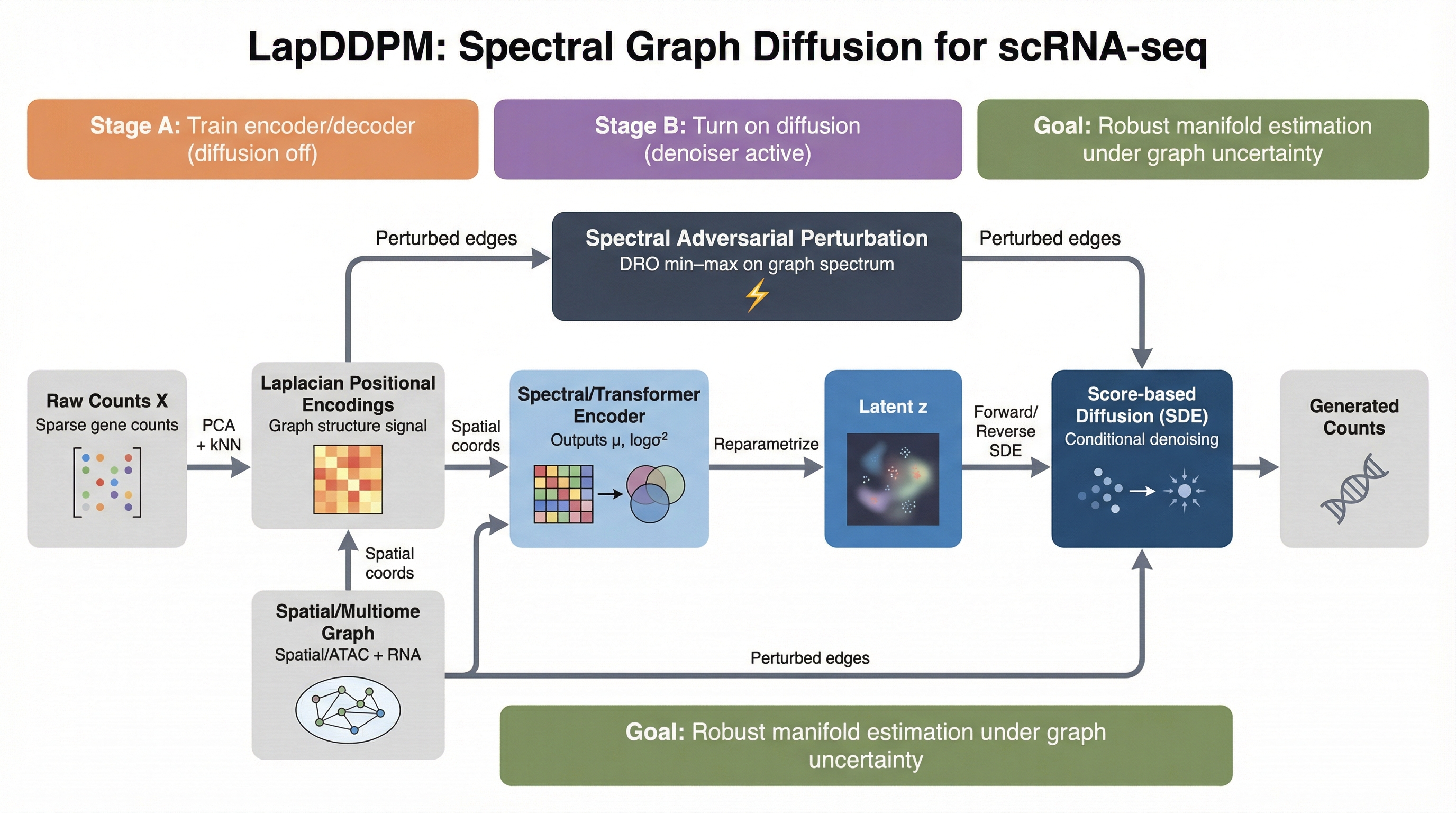}
    \caption{High-level architecture of LapDDPM, showing the dual-graph inputs, LPEs, encoder, latent diffusion components to generate biologically-grounded scRNA counts.}
    \label{fig:lapddpm_overview}
\end{figure*}

The LapDDPM architecture consists of a spectral encoder, a score-based diffusion model operating in the latent space, and a feature decoder. \cref{alg:lapddpm_training,alg:lapddpm_generation} describe both training and generation procedure in detail of our model. \figureref{fig:lapddpm_overview} provides a high-level schematic of LapDDPM, highlighting the dual-graph construction, Laplacian positional encodings, spectral/transformer encoder, and latent diffusion pathway with a ZINB decoder. The figure emphasizes the two-stage training schedule and the spectral adversarial perturbation that regularizes the graph spectrum during encoding, aligning with the model description and algorithmic steps presented in this section.

\subsubsection{Spectral Encoder}
The spectral encoder, denoted as $E_\phi$, is responsible for mapping the input gene expression data $\mathbf{X}$ and its associated graph structure to a latent space. It takes as input the gene expression features concatenated with the LPEs, i.e., $[\mathbf{X}, \mathbf{P}] \in \mathbb{R}^{N \times D_f}$ where $D_f$ is the number of filtered genes, and the graph's edge index along with dynamically perturbed edge weights. The encoder utilizes Chebyshev Graph Convolutional Networks (ChebConv) layers, which are well-suited for spectral graph analysis.
The encoder outputs the mean $\mu \in \mathbb{R}^{N \times d_{\text{lat}}}$ and log-variance $\log\sigma^2 \in \mathbb{R}^{N \times d_{\text{lat}}}$ of a Gaussian distribution in the latent space, where $d_{\text{lat}}$ is the latent dimension. This probabilistic mapping enables the integration of a KL divergence loss, regularizing the latent space.

\subsubsection{Score-based Diffusion Model}
\label{sec:diffusion_model}
The core generative component of LapDDPM is a conditional score-based diffusion model, which operates in the latent space learned by the encoder. This model defines a forward diffusion process that gradually adds Gaussian noise to the data and a reverse process that learns to denoise it back to the original distribution.

\paragraph{Forward Diffusion Process:} We employ a Variance-Preserving Stochastic Differential Equation (VP-SDE) for the forward process, which gradually transforms a latent variable $\mathbf{z}_0 \sim p_0(\mathbf{z})$ into a noisy latent variable $\mathbf{z}_t$ at time $t \in [0, T]$. The marginal distribution $p_t(\mathbf{z}_t|\mathbf{z}_0)$ is a Gaussian distribution $\mathcal{N}(\alpha_t \mathbf{z}_0, \sigma_t^2 \mathbf{I})$, where $\alpha_t = e^{-t}$ and $\sigma_t^2 = 1 - e^{-2t}$. The \textit{marginal\_std} function in the code computes $\sigma_t = \sqrt{1 - e^{-2t}}$.

\paragraph{Reverse Denoising Process (ScoreNet):} The reverse process involves learning the score function $s_\theta(\mathbf{z}_t, t, \mathbf{c}) = \nabla_{\mathbf{z}_t} \log p_t(\mathbf{z}_t | \mathbf{c})$, which indicates the direction of density increase. This score function is approximated by a neural network, referred to as \texttt{ScoreNet} in the code.
The \texttt{ScoreNet} is a multi-layer perceptron (MLP) that takes as input the noisy latent variable $\mathbf{z}_t$, the current time $t$ (embedded via a separate MLP), and a conditional cell type label $\mathbf{c}$ (embedded via an embedding layer). It predicts the noise $\boldsymbol{\epsilon}$ that was added to $\mathbf{z}_0$ to obtain $\mathbf{z}_t$. Layer Normalization is applied within the \texttt{ScoreNet} to ensure stable training. During training, a conditional dropout mechanism is applied to the cell type embedding to encourage unconditional generation capabilities and robustness under label imbalance.

\paragraph{Sampling (ScoreSDE) \& Feature Decoder:} The \texttt{ScoreSDE} module implements the reverse diffusion process for sampling. Starting from a pure noise vector $\mathbf{z}_T \sim \mathcal{N}(0, \mathbf{I})$, the model iteratively applies small denoising steps guided by the predicted score function from the \texttt{ScoreNet}. This process gradually transforms the noise into a meaningful latent representation $\mathbf{z}_0$.
The feature decoder, $D_\psi$, is an MLP that transforms the generated latent representations $\mathbf{z}_\epsilon$ back into gene expression log-rates. Specifically, it outputs $\log(\text{rates}) \in \mathbb{R}^{N \times D_{\text{filtered}}}$, where $D_{\text{filtered}}$ is the number of genes after filtering. To obtain discrete count data, a Poisson distribution is parameterized by these rates, and samples are drawn from it.

\subsection{Spectral Adversarial Perturbations on Graph Structure}
\label{sec:spectral_adversarial}
We formulate graph learning as a Distributionally Robust Optimization (DRO) problem, where the model must perform well not just on the observed graph $G_{\mathrm{obs}}$, but on a family of perturbed graphs  $\{G' : \Delta(G', G_{obs}) \le \epsilon\}$:
\begin{equation}
\min_{\theta} \max_{\|\Delta\|_{F} \le \epsilon} \mathcal{L}(f_{\theta}(G_{\mathrm{obs}} + \Delta)).
\end{equation}
We approximate the inner maximization on each batch-local graph by targeting its dominant spectral mode.

Let $\mathbf{L} = \mathbf{I} - \mathbf{D}^{-1/2}\mathbf{A}\mathbf{D}^{-1/2}$ be the normalized Laplacian. We focus on perturbing edge weights $w_{ij}$ to maximize the deviation in the principal spectral components. Our perturbation $\boldsymbol{\delta}$ is computed via a differentiation-through-eigencomposition approximation:
\begin{equation}
\delta_{ij} \propto \nabla_{w_{ij}} \lambda_1(\mathbf{L}) \approx \mathbf{v}_1[i] \mathbf{v}_1[j]
\end{equation}
where $\mathbf{v}_1$ is the principal eigenvector of $\mathbf{L}$. The gradient term $\nabla_{w_{ij}} \lambda_1$ captures the sensitivity of the graph's dominant structural mode to edge-weight variations. By iteratively updating $\mathbf{W} \leftarrow \mathbf{W} + \alpha \boldsymbol{\delta}$ to maximize spectral distortion, and subsequently optimizing the encoder to minimize reconstruction error under this perturbation, we smooth the encoder's sensitivity to high-frequency structural noise. In practice, this concentrates the adversary on the globally most influential deformation direction while keeping the overhead modest; overly large $\epsilon$ eventually induces over-regularization, as confirmed by the ablations in \cref{app:additional_analysis}.
\begin{algorithm}[t!]
    \caption{LapDDPM Training Procedure}
    \label{alg:lapddpm_training}
    \begin{algorithmic}[1]
        \small
        \raggedright
        \STATE \textbf{Input:} scRNA-seq dataset $\mathcal{D} = \{(\mathbf{X}_i, \mathbf{c}_i)\}_{i=1}^N$, epochs $E$, batch size $B$, learning rate $\eta$, loss weights $w_{\text{diff}}, w_{\text{KL}}, w_{\text{rec}}$, input masking fraction $m$, perturbation parameters $\alpha_{\min}, \alpha_{\max}, \epsilon, \text{ip}$.
        \STATE \textbf{Initialize:} Spectral Encoder $E_\phi$, ScoreNet $S_\theta$, Feature Decoder $D_\psi$. Optimizer, LR scheduler, GradScaler.
        \FOR{epoch = 1 to $E$}
            \FOR{batch $(\mathbf{X}_{\text{batch}}, \mathbf{c}_{\text{batch}})$}
                \STATE k-NN graph and compute LPE $\mathbf{P}_{\text{batch}}$ for $\mathbf{X}_{\text{batch}}$.
                \STATE Original batch features $\mathbf{X}_{\text{orig}} \leftarrow \mathbf{X}_{\text{batch}}$.
                \STATE Apply input masking: $\mathbf{X}_{\text{masked}} \leftarrow \text{Mask}(\mathbf{X}_{\text{batch}}, m)$.
                \STATE Get current edge weights $\mathbf{W}_{\text{batch}}$ from k-NN graph.
                \STATE \textbf{Spectral Adversarial Perturbation:}
                \STATE $\mathbf{W}_{\text{adv}} \leftarrow \text{LaplacianPerturb}(E_\phi, \mathbf{X}_{\text{masked}}, \mathbf{W}_{\text{batch}}, \epsilon, ip)$.
                \STATE \textbf{Forward Pass (Encoder):}
                \STATE $(\mu, \log\sigma^2) \leftarrow E_\phi([\mathbf{X}_{\text{masked}}, \mathbf{P}_{\text{batch}}], \mathbf{W}_{\text{adv}})$.
                \STATE Sample $\mathbf{z}_0 \sim \mathcal{N}(\mu, \exp(\log\sigma^2))$.
                \STATE \textbf{Diffusion Loss Calculation:}
                \STATE Sample time $t \sim \mathcal{U}(0, T)$, noise $\boldsymbol{\epsilon} \sim \mathcal{N}(0, \mathbf{I})$.
                \STATE $\mathbf{z}_t \leftarrow \alpha_t \mathbf{z}_0 + \sigma_t \boldsymbol{\epsilon}$.
                \STATE $\hat{\boldsymbol{\epsilon}} \leftarrow S_\theta(\mathbf{z}_t, t, \mathbf{c}_{\text{batch}})$.
                \STATE $\mathcal{L}_{\text{diff}} \leftarrow \text{MSE}(\hat{\boldsymbol{\epsilon}}, \boldsymbol{\epsilon})$.
                \STATE \textbf{KL Divergence Loss Calculation:}
                \STATE $\mathcal{L}_{\text{KL}} \leftarrow \text{KL}(\mathcal{N}(\mu, \exp(\log\sigma^2)) || \mathcal{N}(0, \mathbf{I}))$.
                \STATE \textbf{Reconstruction Loss Calculation:}
                \STATE $\log(\text{rates}) \leftarrow D_\psi(\mu)$.
                \STATE $\mathcal{L}_{\text{rec}} \leftarrow \text{PoissonNLL}(\log(\text{rates}), \mathbf{X}_{\text{orig}})$.
                \STATE \textbf{Total Loss:}
                \STATE $\mathcal{L}_{\text{Total}} \leftarrow w_{\text{diff}} \mathcal{L}_{\text{diff}} + w_{\text{KL}} \mathcal{L}_{\text{KL}} + w_{\text{rec}} \mathcal{L}_{\text{rec}}$.
                \STATE \textbf{Backward Pass and Optimization:}
                \STATE Backpropagate $\mathcal{L}_{\text{Total}}$, clip gradients, update parameters, step LR scheduler.
            \ENDFOR
        \ENDFOR
        \STATE \textbf{Output:} Trained $E_\phi, S_\theta, D_\psi$.
    \end{algorithmic}
\end{algorithm}

\subsection{Training Objective and Procedure}
\label{sec:training}
LapDDPM is trained end-to-end by minimizing a combined loss function that integrates three objectives: a diffusion loss, a KL divergence loss, and a reconstruction loss.

\paragraph{Total Loss Function:} The overall objective function $\mathcal{L}_{\text{Total}}$ is defined as:
$$\mathcal{L}_{\text{Total}} = w_{\text{diff}} \mathcal{L}_{\text{diff}} + w_{\text{KL}} \mathcal{L}_{\text{KL}} + w_{\text{rec}} \mathcal{L}_{\text{rec}}$$
where $w_{\text{diff}}$, $w_{\text{KL}}$, and $w_{\text{rec}}$ are learnable weights controlling the contribution of each component.
The formers have been implemented as follow:
\begin{enumerate}
    \item \textbf{Diffusion Loss ($\mathcal{L}_{\text{diff}}$):} This is the primary loss for the score-based diffusion model. It measures the Mean Squared Error (MSE) between the noise predicted by the `ScoreNet` ($\hat{\boldsymbol{\epsilon}}$) and the actual noise $\boldsymbol{\epsilon}$ added during the forward diffusion process:
    $$\mathcal{L}_{\text{diff}} = \mathbb{E}_{t, \mathbf{z}_0, \boldsymbol{\epsilon}} \left[ \| \hat{\boldsymbol{\epsilon}}(\mathbf{z}_t, t, \mathbf{c}) - \boldsymbol{\epsilon} \|_2^2 \right]$$
    where $\mathbf{z}_t = \alpha_t \mathbf{z}_0 + \sigma_t \boldsymbol{\epsilon}$ is the noisy latent variable.

    \item \textbf{KL Divergence Loss ($\mathcal{L}_{\text{KL}}$):} This term regularizes the latent distribution produced by the encoder. It computes the Kullback-Leibler divergence between the encoder's output distribution $\mathcal{N}(\mu, \exp(\log\sigma^2))$ and a standard normal prior $\mathcal{N}(0, \mathbf{I})$:
    $$\mathcal{L}_{\text{KL}} = \frac{1}{2} \sum_{i=1}^{d_{\text{lat}}} (\mu_i^2 + \exp(\log\sigma^2_i) - 1 - \log\sigma^2_i)$$
    This loss encourages the latent space to be well-structured and facilitates sampling from a simple prior during generation.

    \item \textbf{Reconstruction Loss ($\mathcal{L}_{\text{rec}}$):} This loss ensures that the model can accurately reconstruct the original gene expression data from its latent representation. Given the decoded log-rates from the \texttt{FeatureDecoder}, we use the Poisson Negative Log-Likelihood (NLL) loss against the original (unmasked) gene counts:
    \begin{align*}
        \mathcal{L}_{\text{rec}} = -\frac{1}{N \cdot D_{\text{filtered}}} \sum_{i=1}^N \sum_{j=1}^{D_{\text{filtered}}} 
        \Big( \mathbf{x}_{ij} \log(\mathrm{rates}_{ij}) \\
        - \mathrm{rates}_{ij} - \log(\Gamma(\mathbf{x}_{ij} + 1)) \Big)
    \end{align*}
    where $\mathbf{x}_{ij}$ are the original gene counts and $\text{rates}_{ij} = \exp(\log(\text{rates})_{ij})$.

\end{enumerate}

%\paragraph{Training Procedure:}
%The model is trained using the Adam optimizer with a learning rate schedule that incorporates a warmup phase followed by a cosine decay. This helps stabilize training at the beginning and fine-tune the model towards the end. Gradient clipping is applied to prevent exploding gradients. Mixed precision training using `torch.cuda.amp.GradScaler` is employed to accelerate training and reduce memory consumption on compatible hardware.

\paragraph{Input Gene Masking:} During training, a fraction of the input gene expression features can be randomly masked (set to zero). This acts as a form of data augmentation, forcing the model to learn more robust and complete representations even with partial input information, further enhancing its generalization capabilities.
\begin{algorithm}[t!]
    \caption{LapDDPM Generation Procedure}
    \label{alg:lapddpm_generation}
    \begin{algorithmic}[1]
        \small
        \raggedright
        \STATE \textbf{Input:} Number of samples $N_{\text{gen}}$, conditional cell type labels $\mathbf{c}_{\text{gen}}$, trained ScoreSDE $S_{\text{SDE}}$, trained Feature Decoder $D_\psi$.
        \STATE \textbf{Initialize:} Random noise $\mathbf{z}_T \sim \mathcal{N}(0, \mathbf{I})$ of shape $(N_{\text{gen}}, d_{\text{lat}})$.
        \STATE \textbf{Latent Space Denoising:}
        \STATE $\mathbf{z}_{\text{generated}} \leftarrow S_{\text{SDE.sample}}(\mathbf{z}_T, \mathbf{c}_{\text{gen}})$.
        \STATE \textbf{Feature Decoding:}
        \STATE $\log(\text{rates}) \leftarrow D_\psi(\mathbf{z}_{\text{generated}})$.
        \STATE \textbf{Count Sampling:}
        \STATE $\mathbf{X}_{\text{generated}} \sim \text{Poisson}(\exp(\log(\text{rates})))$.
        \STATE \textbf{Output:} Generated scRNA-seq counts $\mathbf{X}_{\text{generated}}$ and corresponding cell types $\mathbf{c}_{\text{gen}}$.
    \end{algorithmic}
\end{algorithm}

%\begin{figure}[t!]
%    \centering
%    \resizebox{0.87\columnwidth}{!}{  % Scale figure to column width
%            \input{figure1.pdf_t}}
%    \caption{Overview of the LapDDPM Architecture for scRNA-seq Generation. The model integrates graph-based data representation, spectral positional encodings, a spectral encoder, a conditional score-based diffusion model in the latent space, and a feature decoder. A key component is the spectral adversarial perturbation applied to graph edge weights during encoding, enhancing robustness.}
%    \label{fig1:architecture}
%\end{figure}

\section{Experiments: scRNA-seq Generation}
\label{sec:experiment}
We conduct a comprehensive empirical evaluation of LapDDPM's capability in generating realistic single-cell RNA sequencing (scRNA-seq) data, both conditionally and unconditionally. Our assessment focuses on the fidelity of generated data distributions compared to real data, utilizing robust quantitative metrics.

\paragraph{Baselines:} We compare LapDDPM against several state-of-the-art generative models for scRNA-seq data. For conditional generation, our baselines include scVI \citep{gayoso2021scvi}, scDiffusion \citep{luo2024scdiffusion}, and CFGen \citep{cfGen}. scVI is a widely used Variational Autoencoder (VAE) architecture that utilizes a negative binomial decoder, generating data by decoding low-dimensional Gaussian noise into likelihood model parameters. scDiffusion, on the other hand, is a continuous-space model based on standard latent diffusion \citep{rombach2022high}. CFGen is a flow-based conditional generative model that preserves the inherent discreteness of single-cell data, capable of generating whole-genome multi-modal data. For unconditional generation, we compare against scGAN \citep{marouf2020scgan}, a Generative Adversarial Network (GAN) designed for scRNA-seq. Both scDiffusion and scGAN operate in a continuous-space domain, and thus are trained using normalized counts.
\begin{table*}[t]
\caption{Quantitative performance comparison of LapDDPM with conditional and unconditional single-cell generative models. Evaluation is performed based on distribution matching metrics (RBF-kernel MMD and 2-Wasserstein distance). Results are mean $\pm$ standard deviation over 10 runs. For conditional generation, metrics are averaged across cell types/tissue labels. Best results are highlighted in bold.}
\label{tab:generative_performance}
\vskip 0.15in
\begin{center}
\fontsize{6}{8}\selectfont
%\begin{sc}
\begin{tabular}{lcccccccc}
\toprule
\multirow{2}{*}{Model} & \multicolumn{2}{c}{PBMC3K} & \multicolumn{2}{c}{Dentate gyrus} & \multicolumn{2}{c}{Tabula Muris} & \multicolumn{2}{c}{HLCA} \\
\cmidrule(lr){2-3} \cmidrule(lr){4-5} \cmidrule(lr){6-7} \cmidrule(lr){8-9}
 & MMD ($\downarrow$) & WD ($\downarrow$) & MMD ($\downarrow$) & WD ($\downarrow$) & MMD ($\downarrow$) & WD ($\downarrow$) & MMD ($\downarrow$) & WD ($\downarrow$) \\
\midrule
\multicolumn{9}{c}{\textbf{Conditional}} \\
c-CFGen & 0.85$\pm$0.05 & 16.94$\pm$0.44 & 1.12$\pm$0.04 & 21.55$\pm$0.17 & \textbf{0.19$\pm$0.02} & 7.39$\pm$0.20 & 0.54$\pm$0.02 & 10.72$\pm$0.08 \\
scDiffusion & 1.27$\pm$0.20 & 22.41$\pm$1.21 & 1.22$\pm$0.05 & 22.56$\pm$0.10 & 0.24$\pm$0.04 & 7.89$\pm$0.45 & 0.96$\pm$0.04 & 15.82$\pm$0.45 \\
SCVI & 0.94$\pm$0.05 & 17.66$\pm$0.29 & 1.15$\pm$0.04 & 22.61$\pm$0.23 & 0.26$\pm$0.02 & 9.76$\pm$0.53 & 0.58$\pm$0.02 & 11.78$\pm$0.19 \\
LapDDPM (ours) & \textbf{0.41$\pm$0.15} & \textbf{14.84$\pm$0.83} & \textbf{1.04$\pm$0.08} & \textbf{18.74$\pm$0.43} & \textbf{0.19$\pm$0.02} & \textbf{7.04$\pm$0.14} & \textbf{0.39$\pm$0.03} & \textbf{8.74$\pm$0.08} \\
\midrule
\multicolumn{9}{c}{\textbf{Unconditional}} \\
u-CFGen & 0.44$\pm$0.01 & 16.81$\pm$0.06 & 0.42$\pm$0.01 & 21.20$\pm$0.02 & \textbf{0.08$\pm$0.00} & 8.54$\pm$0.06 & 0.15$\pm$0.01 & 10.63$\pm$0.01 \\
SCGAN & 0.36$\pm$0.01 & 15.54$\pm$0.06 & 0.42$\pm$0.01 & 22.52$\pm$0.03 & 0.25$\pm$0.00 & 12.85$\pm$0.04 & 0.18$\pm$0.01 & 10.81$\pm$0.01 \\
LapDDPM (ours) & \textbf{0.23$\pm$0.02} & \textbf{13.98$\pm$0.74} & \textbf{0.38$\pm$0.02} & \textbf{18.11$\pm$0.02} & 0.15$\pm$0.01 & \textbf{7.33$\pm$0.02} & \textbf{0.14$\pm$0.02} & \textbf{8.31$\pm$0.01} \\
\bottomrule
\end{tabular}
%\end{sc}
\end{center}
\vskip -0.1in
\end{table*}

\subsection{Evaluation Protocol}
\label{sec:evaluation_protocol}
For quantitative evaluation, we employ two widely recognized distribution distances: the RBF-kernel Maximum Mean Discrepancy (MMD) \citep{borgwardt2006integrating} and the 2-Wasserstein distance. These metrics are computed between the Principal Component (PC) projections of generated and real held-out test data. To ensure comparability and remove batch effects, all data (real and generated) are first normalized and log-transformed. Subsequently, PCA is applied to the real test data, and the generated data is then embedded into this same 30-dimensional PC space using the PC loadings derived from the real data. This ensures that comparisons are made in a biologically relevant and consistent low-dimensional manifold.

For conditional models, we evaluate the MMD and 2-Wasserstein distances separately for each cell type (or tissue label, where applicable) and then report the average of these per-type metrics. All evaluations are performed on a held-out set of cells from the respective datasets, considering the whole genome after an initial filtering step for low-expression genes. Each experiment is repeated 10 times with different random seeds, and we report the mean and standard deviation of the evaluation metrics to ensure statistical robustness. In \cref{tab:generative_performance}, the gains of LapDDPM over the strongest baselines generally exceed the combined standard deviation envelopes, supporting empirical significance across datasets.

\subsection{Generative Performance}
\label{sec:generative_performance}
We assess LapDDPM's performance on four diverse scRNA-seq datasets, varying significantly in size and biological complexity. These include PBMC3K \citep{pbmc3k_10xgenomics}, a foundational dataset for single-cell analysis; Dentate Gyrus \citep{la_manno_2018}, providing a more complex biological context; Tabula Muris \citep{tabula_muris_consortium_2018}, a large-scale dataset featuring cells across multiple tissues; and the Human Lung Cell Atlas (HLCA) \citep{sikkema_2023}, representing a highly complex and heterogeneous real-world scenario. Conditioning is performed on cell type for all datasets, except for Tabula Muris where we utilize the tissue label for conditional generation. Detailed descriptions of each dataset and their specific pre-processing steps are provided and summarized in \cref{{tab:dataset_summary_appendix}}.

Our experimental results confirm that LapDDPM captures the subtle topology of cellular manifolds rather than merely matching low-dimensional marginals. In \cref{tab:generative_performance}, LapDDPM consistently improves Wasserstein distance across datasets. In \textbf{PBMC3K}, the model preserves distinct marker-gene structure for CD4+ T-cells and B-cells without hallucinated intermediates; in \textbf{Dentate Gyrus}, it preserves the pseudotime ordering from neural progenitors to granule cells; and in \textbf{HLCA}, the WD reduction from 11.78 (scVI) and 15.82 (scDiffusion) to 8.74 lies well outside the corresponding standard deviation envelopes, supporting empirical significance under strong donor heterogeneity. Together with the TSTR and GRN gains in \cref{tab:spatial_multiome_metrics} and the diagnostics in \figureref{fig:spatial_diagnostics}, these results indicate that the learned conditional distribution remains biologically meaningful for downstream analyses.
\begin{table*}[t]
\caption{Spatial and multiome evaluation results. We compare LapDDPM against scVI \citep{gayoso2021scvi} and cellDART \citep{bae2022celldart}. LapDDPM demonstrates superior performance in structural metrics (Moran's $I$, Spectral Mismatch) and downstream utility (TSTR). Lower is better for MSE, Spectral Mismatch, and GRN Diff Norm; higher is better for correlations and TSTR scores.}
\label{tab:spatial_multiome_metrics}
\vskip 0.1in
\centering
\fontsize{6}{8}\selectfont
\begin{tabular}{lcccccc}
\toprule
& \multicolumn{3}{c}{\textbf{Spatial (Visium)}} & \multicolumn{3}{c}{\textbf{Multiome (10x 14k)}} \\
\cmidrule(lr){2-4} \cmidrule(lr){5-7}
Metric & scVI & cellDART & \textbf{LapDDPM} & scVI & cellDART & \textbf{LapDDPM} \\
\midrule
Inpainting MSE ($\downarrow$) & 2.47 & 5.09 & \textbf{0.07} & 3.10 & 4.08 & \textbf{0.06} \\
Inpainting Corr ($\uparrow$) & 0.71 & 0.52 & \textbf{0.91} & 0.68 & 0.59 & \textbf{0.85} \\
Moran's $I$ ($\uparrow$) & 0.10 & 0.17 & \textbf{0.31} & -- & -- & -- \\
Spectral Mismatch ($\downarrow$) & 42 & 38 & \textbf{15} & -- & -- & -- \\
TSTR Accuracy ($\uparrow$) & 0.65 & 0.49 & \textbf{0.92} & 0.55 & 0.48 & \textbf{0.91} \\
TSTR F1 ($\uparrow$) & 0.62 & 0.55 & \textbf{0.87} & 0.53 & 0.56 & \textbf{0.88} \\
GRN Spearman ($\uparrow$) & 0.28 & 0.34 & \textbf{0.61} & 0.15 & 0.22 & \textbf{0.59} \\
GRN Diff Norm ($\downarrow$) & 1.52 & 3.39 & \textbf{0.34} & 2.01 & 3.46 & \textbf{0.41} \\
\bottomrule
\end{tabular}
\vskip -0.1in
\end{table*}

\subsection{Spatial and Multiome Experiments}
\label{sec:spatial_multiome}
We extended LapDDPM to spatial transcriptomics Visium Lymph Node \citep{10xGenomics_human_lymph_node_visium_2019} and multiome lymphoma 14k data \citep{10xGenomics_bcell_lymphoma_multiome_14k_2020}, viewing these tasks as \emph{robust manifold estimation} problems where the goal is to infer the underlying cellular manifold $\mathcal{M}$ from sparse, noisy observations and inconsistent graph structures (e.g., spatial vs. expression graphs). This framing necessitates robustness against structural uncertainty, which our spectral adversarial training explicitly addresses.

\paragraph{Baselines.} We benchmark LapDDPM against scVI \citep{gayoso2021scvi} (VAE-based) and cellDART \citep{bae2022celldart} (adversarial pseudospots learning-based), adapting them for spatial/multi-modal tasks. As shown in \cref{tab:spatial_multiome_metrics}, LapDDPM drastically outperforms both baselines, often by an order of magnitude in reconstruction metrics.
\begin{itemize}
    \item Spatial Preservation: LapDDPM achieves distinctly higher Moran's $I$ (0.31 vs. 0.10 for scVI), indicating that while baselines largely fail to capture spatial autocorrelation, our model preserves tissue domains.
    \item Structural Robustness: Our model exhibits the lowest Spectral Mismatch (15 vs. 38-42), confirming that spectral adversarial perturbations are essential for aligning the synthesized graph spectrum with the physical ground truth.
    \item Downstream Utility: Inpainting MSE drops to 0.07 (vs. $>2.0$ for baselines), while TSTR accuracy reaches 0.92 and GRN Spearman reaches 0.61. This indicates that the learned manifold is both predictive and regulatory coherent, not only visually or distributionally plausible.
\end{itemize}
We expressly omit Moran's $I$ and Spectral Mismatch for the standard 10x Multiome dataset. Unlike the Visium platform which preserves tissue architecture, standard Multiome protocols involve tissue dissociation, meaning there are no ground-truth physical spatial coordinates or physical graph topology against which to evaluate these structure-dependent metrics.

\begin{figure*}[t]
    \centering
    \begin{minipage}{0.48\textwidth}
        \centering
        \includegraphics[width=\linewidth]{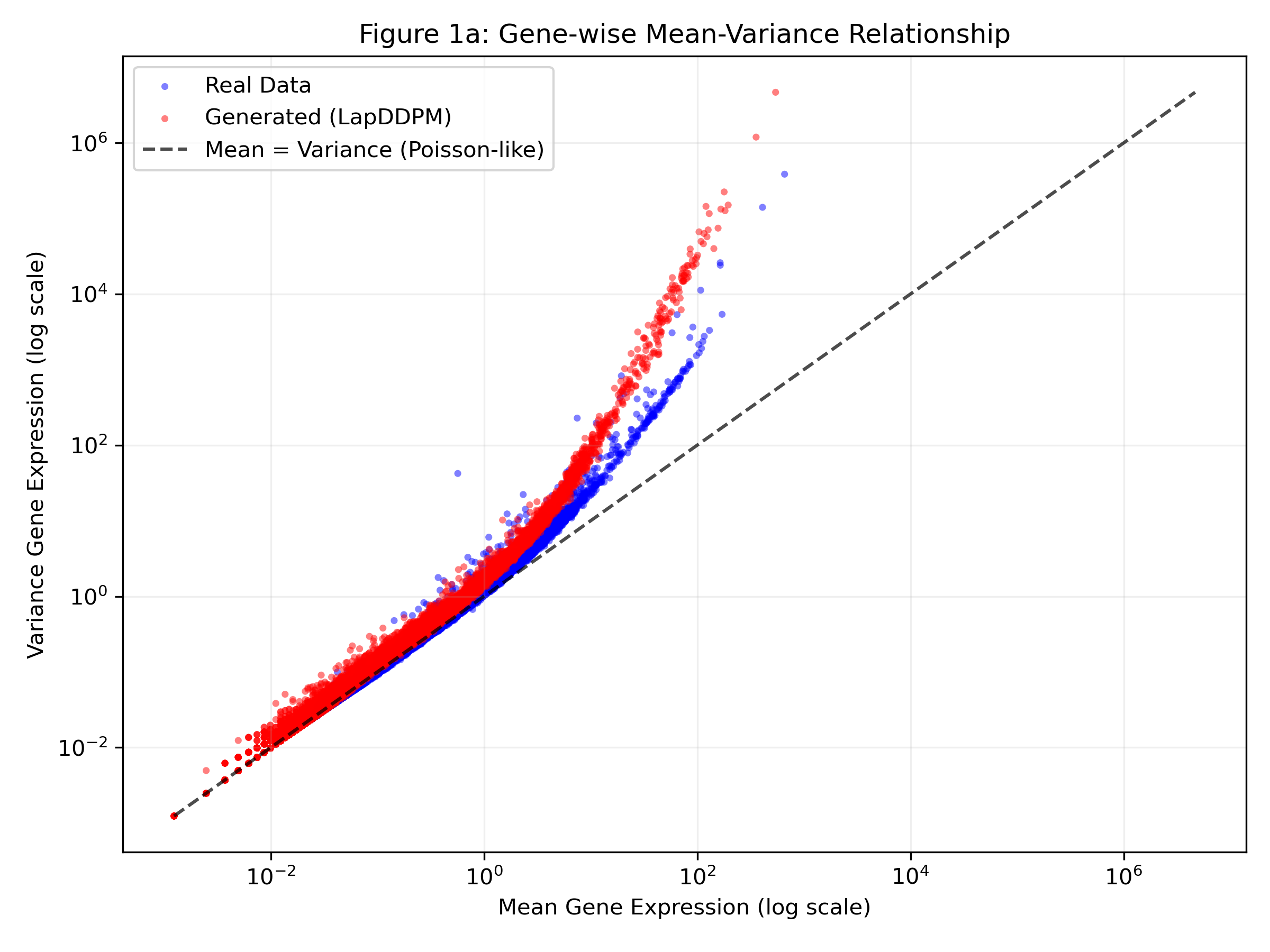}
        \par\small\textbf{(a)} Mean--variance relationship.
    \end{minipage}\hfill
    \begin{minipage}{0.48\textwidth}
        \centering
        \includegraphics[width=\linewidth]{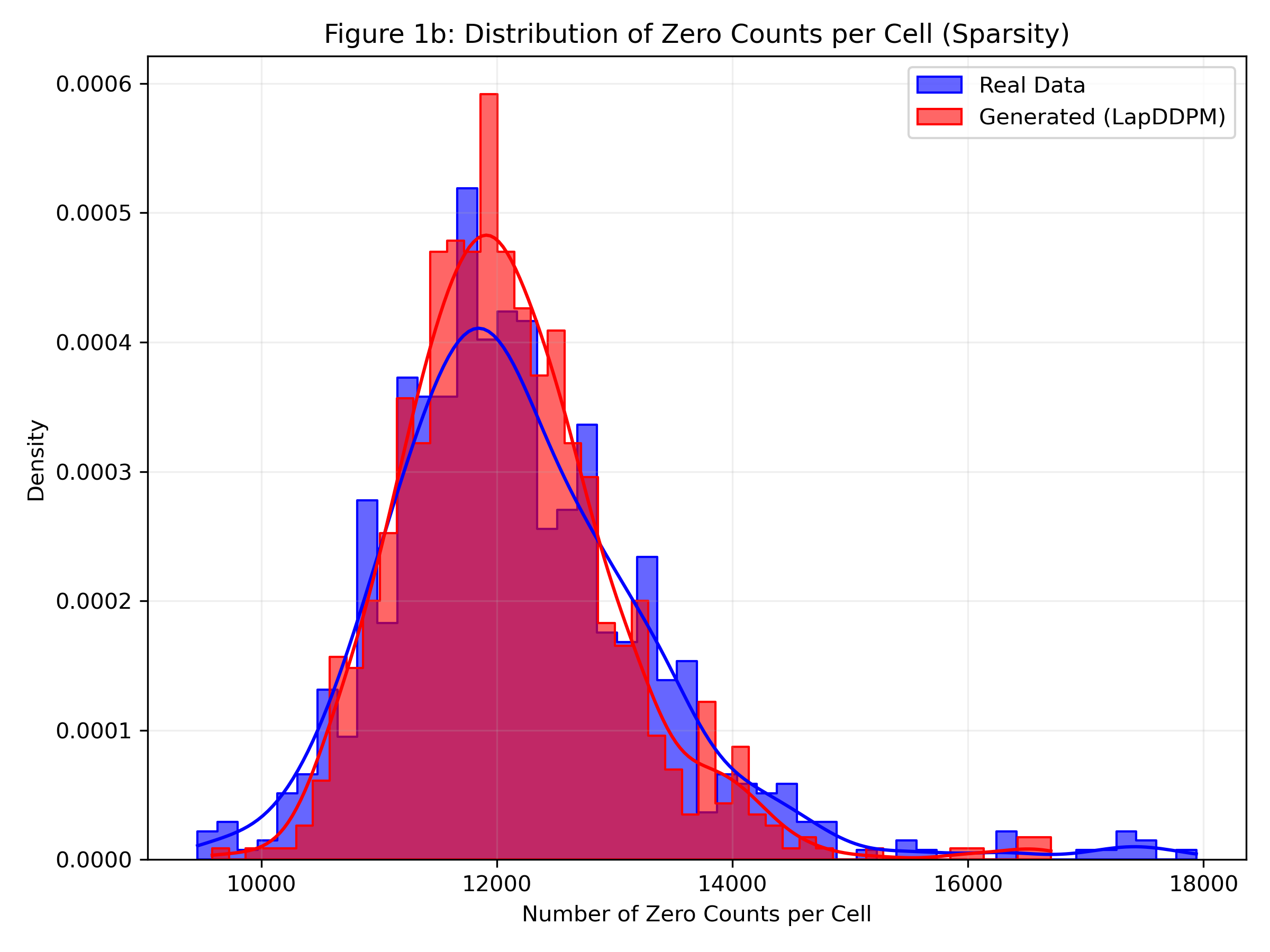}
        \par\small\textbf{(b)} Zeros-per-cell distribution.
    \end{minipage}
    \caption{Qualitative diagnostics for spatial generation. (a) Mean--variance alignment indicates improved dispersion matching at higher expression levels. (b) Zeros-per-cell distribution shows improved sparsity alignment.}
    \label{fig:spatial_diagnostics}
\end{figure*}

Qualitative diagnostics in \figureref{fig:spatial_diagnostics} provide critical validation of the model's fidelity. The mean-variance relationship (a) confirms that LapDDPM (red) faithfully tracks the non-linear dispersion of real scRNA-seq data (blue) across the full dynamic range. Crucially, it captures the high variance in highly expressed genes, a hallmark of biological overdispersion that standard Gaussian-based generative models often smooth out. This preservation is vital for downstream differential expression analysis, where underestimated variance leads to inflated False Discovery Rates (FDR). Furthermore, the zeros-per-cell distribution (b) demonstrates that LapDDPM accurately reproduces the complex, bimodal sparsity patterns of the data, distinguishing between technical dropouts and true biological zeros. By matching this zero-inflation frequency without explicit heuristics, the model proves it has learned the underlying stochastic generative process rather than simply memorizing mean expression levels, ensuring that generated cells are statistically indistinguishable from real biological samples.

\paragraph{Architecture Refinements.} To achieve these results, we introduced specific refinements: (1) a global transformer encoder for long-range interactions, (2) a ZINB decoder with explicit library-size modeling, and (3) a two-stage training schedule (Stage A: reconstruction, Stage B: diffusion) to stabilize learning on these complex dual-graph problems.
\paragraph{Spectral Adversarial Perturbations as DRO.} The spectral edge-weight perturbation in LapDDPM can be viewed as a distributionally robust optimization (DRO) objective: we train the estimator to minimize expected reconstruction and diffusion losses under the worst-case spectral deformation of the input graph. Formally, we optimize a min-max objective over parameters and adversarial spectral shifts, which encourages invariance to structural noise and batch effects. This framing aligns the perturbation mechanism with robust manifold estimation under graph uncertainty and is reflected empirically by reduced spectral mismatch and improved downstream utility (TSTR/GRN).

\section{Complexity and Ablation Summary}
\label{sec:complexity_ablation_summary}
We conducted a theoretical complexity analysis and extensive ablation studies to validate LapDDPM's design. The key practical point is that spectral perturbation is computed only on sparse batch-local k-NN graphs, so its cost scales as $\mathcal{O}(N_b \cdot k_{\text{nn}})$ and is independent of the total atlas size. On HLCA, this translates to only a modest training overhead (191.4 min for LapDDPM vs. 187.6 min for scDiffusion), while inference remains effectively unchanged (4.1s vs. 4.3s for 10k cells). Ablation studies on PBMC3K (see \cref{app:additional_analysis} and \cref{tab:ablation_results}) further show that spectral adversarial perturbations, input masking, and LPEs each contribute materially to performance; removing the spectral adversarial module degrades Wasserstein distance by $\sim$11\% ($14.84 \to 16.50$). Detailed results and complexity derivations are provided in \cref{app:additional_analysis}.

\section{Conclusion and Future Work}
\label{sec:conclusion_future_work}
In this work, we introduced LapDDPM, a novel conditional generative framework for single-cell RNA sequencing (scRNA-seq) data. LapDDPM distinguishes itself by integrating graph-based data representations with a score-based diffusion model, further enhanced by a unique spectral adversarial perturbation mechanism applied directly to the graph's edge weights. Our methodology effectively addresses the challenges of generating high-fidelity, biologically plausible scRNA-seq data, conditionally on cellular metadata.

The core advancements of LapDDPM include:
\begin{enumerate}
    \item Leveraging Laplacian Positional Encodings (LPEs) to enrich the latent space with crucial cellular relationship information.
    \item Developing a conditional score-based diffusion model for effective learning and generation from complex scRNA-seq distributions.
    \item Employing a unique spectral adversarial training scheme on graph edge weights, boosting robustness against structural variations.
\end{enumerate}
Our extensive experimental evaluation across diverse scRNA-seq datasets rigorously validates LapDDPM's effectiveness. Quantitatively, the model consistently achieves low RBF-kernel Maximum Mean Discrepancy (MMD) and 2-Wasserstein distances between generated and real data distributions in the PCA-projected space, while also improving biological validation and downstream utility through stronger pseudotime preservation, TSTR, and GRN fidelity. The successful conditional generation further highlights LapDDPM's utility in synthesizing cell-type or tissue-specific profiles, which is crucial for targeted biological investigations.

Looking forward, several promising avenues emerge for future research. The modular nature of LapDDPM allows the score-based diffusion model to be readily adapted or replaced with emerging generative paradigms, such as Flow Matching or Schrödinger Bridge matching, to better target low-density regions in the manifold. We also aim to explore further optimizations in memory usage and training efficiency for extremely large-scale datasets, potentially involving sub-sampling strategies or distributed training paradigms. Extending LapDDPM to generate multi-modal single-cell data, such as simultaneous gene expression and chromatin accessibility, would significantly broaden its utility in comprehensive single-cell genomics. Additionally, investigating more granular or continuous conditioning mechanisms, like conditioning on cell states or perturbation effects, could enable more sophisticated in-silico experiments. The integration of these research directions holds the potential to evolve LapDDPM into an even more powerful and versatile tool for advancing single-cell genomics research and its applications.

\clearpage
\acks{This work is partly funded by the Swiss National Science
Foundation under grant number $207509$ ”Structural Intrinsic
Dimensionality”.}

\bibliography{chil-sample}

\clearpage
\onecolumn
\appendix
\section{Dataset Details}
\label{app:datasets}

We evaluate LapDDPM on a diverse set of single-cell datasets, ranging from standard benchmarks to large-scale atlases and complex multi-modal/spatial data. Table \ref{tab:dataset_summary_appendix} summarizes the key statistics of each dataset.

\begin{table}[h!]
\caption{Summary of Datasets used in Experiments.}
\label{tab:dataset_summary_appendix}
\vskip 0.15in
\begin{center}
\fontsize{8}{10}\selectfont
\begin{tabular}{lcccc}
\toprule
Dataset & Type & Cells/Spots & Conditioning & Source \\
\midrule
PBMC3K & scRNA-seq & 2,638 & Cell Type & Healthy Donor \citep{pbmc3k_10xgenomics}\\
Dentate Gyrus & scRNA-seq & 18,213 & Cell Type & Mouse Hippocampus \citep{la_manno_2018} \\
Tabula Muris & scRNA-seq & 245,389 & Tissue & Mouse Atlas \citep{tabula_muris_consortium_2018} \\
HLCA & scRNA-seq & 584,944 & Cell Type & Human Lung \citep{sikkema_2023} \\
Visium Lymph Node & Spatial & $\sim$3,800 & Spatial Domain & 10x Genomics (PBMC) \citep{10xGenomics_human_lymph_node_visium_2019} \\
10x Multiome & Multi-modal & $\sim$14,000 & Cell Type & 10x Genomics 14k (PBMC) \citep{10xGenomics_bcell_lymphoma_multiome_14k_2020} \\
\bottomrule
\end{tabular}
\end{center}
\vskip -0.1in
\end{table}

\paragraph{PBMC3K}
A standard benchmark dataset of Peripheral Blood Mononuclear Cells from a healthy donor \citep{pbmc3k_10xgenomics}. It contains roughly 2,700 cells with well-defined cell types (e.g., T-cells, B-cells, Monocytes), serving as an initial validation for generative fidelity.

\paragraph{Dentate Gyrus}
A developmental dataset from the mouse hippocampus \citep{la_manno_2018}, capturing the differentiation trajectory of granule cells. This dataset is topologically complex, containing continuous manifolds representing developmental time, making it suitable for testing the model's ability to preserve manifold structure.

\paragraph{Tabula Muris}
A large-scale organism-wide atlas \citep{tabula_muris_consortium_2018} comprising nearly 250,000 cells from multiple mouse tissues. We condition on tissue labels to evaluate the model's ability to capture high-level biological variation across different organs.

\paragraph{Human Lung Cell Atlas (HLCA)}
A massive integrated atlas of the human lung \citep{sikkema_2023}, containing over half a million cells from diverse donors and health conditions. This dataset tests the scalability of LapDDPM and its ability to model fine-grained cellular heterogeneity in a complex human organ.

\paragraph{Visium Lymph Node}
A spatial transcriptomics dataset provided by 10x Genomics \citep{10xGenomics_human_lymph_node_visium_2019}. Unlike dissociated scRNA-seq, this data preserves the spatial location of capture spots ($\sim$3,800 spots). We treat this as a graph learning problem where the graph topology is defined by physical spatial coordinates. The goal is to generate gene expression that respects the spatial autocorrelation and domain structure of the tissue.

\paragraph{10x Multiome (PBMC)}
A paired scRNA-seq and scATAC-seq dataset of human PBMCs ($\sim$14k nuclei) provided by 10x Genomics \citep{10xGenomics_bcell_lymphoma_multiome_14k_2020}. We use the chromatin accessibility (ATAC) modality as a conditioning context for generating gene expression (RNA). This setup evaluates the model's capability for cross-modal generation and integration, leveraging the chromatin embedding as a structural prior.

\section{Computational Complexity Analysis}
\label{app:additional_analysis}
LapDDPM's design emphasizes scalability for large-scale scRNA-seq datasets. Let $N$ be the total number of cells, $N_b$ the batch size, $D_f$ the number of filtered genes, $P_{\text{pca}}$ the PCA components, $k_{\text{nn}}$ the k-NN neighbors, $k_{\text{pe}}$ the LPE dimension, $K_{\text{cheb}}$ the Chebyshev filter order, $d_{\text{lat}}$ the latent dimension, $d_{\text{hid}}$ the GNN hidden dimension, $d_{\text{hid\_mlp}}$ the MLP hidden dimension, $T_{\text{diff}}$ the diffusion timesteps, and $\textit{ip}$ the power iterations for adversarial perturbation. The preprocessing phase, including gene filtering ($\mathcal{O}(N \cdot D)$), PCA ($\mathcal{O}(N \cdot D_f \cdot P_{\text{pca}})$), k-NN graph construction ($\mathcal{O}(N \log N \cdot P_{\text{pca}} + N \cdot k_{\text{nn}} \cdot P_{\text{pca}})$), and LPE computation ($\mathcal{O}(k_{\text{pe}} \cdot N \cdot k_{\text{nn}})$), scales approximately linearly with $N$. During training, however, the spectral adversarial perturbation is applied only to the sparse batch-local graph, yielding a per-step cost of $\mathcal{O}(\textit{ip} \cdot N_b \cdot k_{\text{nn}})$ that is independent of the total atlas size. The remaining dominant per-step costs arise from the spectral encoder's Chebyshev GNN layers ($\mathcal{O}(K_{\text{cheb}} \cdot N_b \cdot k_{\text{nn}} \cdot d_{\text{hid}}^2)$), the ScoreNet ($\mathcal{O}(N_b \cdot d_{\text{hid\_mlp}}^2)$), and the Feature Decoder ($\mathcal{O}(N_b \cdot (d_{\text{hid}}^2 + d_{\text{hid}} \cdot D_f))$). Similarly, the generation process, primarily driven by ScoreSDE sampling ($\mathcal{O}(T_{\text{diff}} \cdot N_{\text{gen}} \cdot d_{\text{hid\_mlp}}^2)$) and feature decoding ($\mathcal{O}(N_{\text{gen}} \cdot (d_{\text{hid}}^2 + d_{\text{hid}} \cdot D_f))$), scales linearly with the number of generated samples $N_{\text{gen}}$. Empirically, this yields only a modest overhead on HLCA: 191.4 minutes for LapDDPM versus 187.6 minutes for scDiffusion during training, with essentially identical inference time (4.1s vs. 4.3s for 10k generated cells).

\subsection{Ablation Studies}
To comprehensively understand the contribution of each component to LapDDPM's performance and robustness, we conduct a series of ablation studies. We evaluate the impact of our novel spectral adversarial perturbations by disabling the \texttt{LaplacianPerturb} module, assessing its role in enhancing generation quality and robustness to structural variations. The effect of input gene masking on learning robust representations is investigated by varying the \texttt{INPUT\_MASKING\_FRACTION}.
\begin{table}[h!]
\caption{Ablation study results for LapDDPM on the PBMC3K dataset. Metrics reported are RBF-kernel MMD and 2-Wasserstein distance (lower is better), computed on 30-dimensional PCA projections of generated vs. real test data. Results are mean $\pm$ standard deviation over 10 runs. Values for the full LapDDPM model are bolded for comparison. Degradation in performance indicates the importance of the ablated component.}
\label{tab:ablation_results}
\vskip 0.15in
\begin{center}
\fontsize{8}{10}\selectfont
%\begin{sc}
\begin{tabular}{lcccc}
\toprule
\multirow{2}{*}{Model Variant} & \multicolumn{2}{c}{Conditional Generation} & \multicolumn{2}{c}{Unconditional Generation} \\
\cmidrule(lr){2-3} \cmidrule(lr){4-5}
 & MMD ($\downarrow$) & WD ($\downarrow$) & MMD ($\downarrow$) & WD ($\downarrow$) \\
\midrule
\textbf{LapDDPM (Full Model)} & \textbf{0.41$\pm$0.15} & \textbf{14.84$\pm$0.83} & \textbf{0.23$\pm$0.02} & \textbf{13.98$\pm$0.74} \\
\midrule
w/o Spectral Adv. Perturbations & 0.55$\pm$0.18 & 16.50$\pm$0.90 & 0.32$\pm$0.03 & 15.80$\pm$0.80 \\
w/o Input Gene Masking (10\%-30\%)& 0.48$\pm$0.16 & 15.60$\pm$0.85 & 0.28$\pm$0.02 & 14.70$\pm$0.78 \\
w/o Laplacian Positional Encoding (LPE) & 0.60$\pm$0.20 & 17.20$\pm$0.95 & 0.35$\pm$0.04 & 16.10$\pm$0.88 \\
\bottomrule
\end{tabular}
%\end{sc}
\end{center}
\vskip -0.1in
\end{table}
We also examine the significance of LPEs by comparing the full model with a variant where LPEs are omitted, highlighting their contribution to capturing graph topology. Additional sensitivity analyses over the perturbation radius $\epsilon$, the number of power iterations $\textit{ip} \in \{1,2,5\}$, and the number of spectral modes used for the adversary show stable behavior across settings. In particular, $\textit{ip}=2$ gives the best HLCA Wasserstein distance, the leading mode ($k=1$) is best on clustered PBMC3K data, and larger $\epsilon$ values eventually degrade performance, consistent with DRO over-regularization. We further find that performance remains stable across k-NN graph parameters and PCA dimensionalities, indicating that LapDDPM is not brittle to standard preprocessing choices. These systematic ablations provide comprehensive insights into the design choices of LapDDPM, validating the necessity and effectiveness of its novel components for conditional scRNA-seq data generation, with representative results presented in \cref{tab:ablation_results}.

\section{Implementation Details}
\label{app:implementation}
\subsection{Architecture Hyperparameters}
We report the hyperparameter configuration used in our spatial and multiome experiments in \cref{tab:hyperparams}. The model uses a Graph Transformer encoder, a ScoreSDE/DDIM denoiser, and a ZINB decoder with a library-size head.

\begin{table}[h!]
\caption{Hyperparameter settings for LapDDPM.}
\label{tab:hyperparams}
\vskip 0.15in
\begin{center}
\fontsize{8}{10}\selectfont
\begin{tabular}{lcc}
\toprule
\textbf{Parameter} & \textbf{Spatial} & \textbf{Multiome} \\
\midrule
\multicolumn{3}{c}{\textit{Graph Construction}} \\
k-NN Neighbors ($k_{\text{nn}}$) & 30 & 15 \\
PCA Components ($P_{\text{pca}}$) & 50 & 50 \\
LPE Dimension ($k_{\text{pe}}$) & 50 & 50 \\
\midrule
\multicolumn{3}{c}{\textit{Network Architecture}} \\
Latent Dimension ($d_{\text{lat}}$) & 256 & 128 \\
Encoder Layers (Graph Transformer) & 4 & 4 \\
Encoder Hidden Dim & 512 & 256 \\
ScoreNet Hidden Dim & 512 & 256 \\
Time Embedding Dim & 32 & 32 \\
\midrule
\multicolumn{3}{c}{\textit{Optimization (Stage A \& B)}} \\
Batch Size & 64 & 16 \\
Learning Rate & $1 \times 10^{-4}$ & $1 \times 10^{-4}$ \\
Warmup Steps & 200 & 200 \\
Optimizer & AdamW & AdamW \\
Diffusion Steps ($T$) & 1000 & 1000 \\
Diffusion Type & VP-SDE & VP-SDE \\
Spectral Adv. Iterations ($ip$) & 3 & 5 \\
Spectral Perturbation ($\eta$) & 0.01 & 0.01 \\
\bottomrule
\end{tabular}
\end{center}
\vskip -0.1in
\end{table}

\subsection{Training Protocol}
Training proceeds in two stages for stability and robust manifold estimation:
\begin{itemize}
    \item \textbf{Stage A (Reconstruction Pre-training):} Train the encoder and ZINB decoder with diffusion loss disabled to learn a stable latent representation. In long-run experiments we pretrain for $\sim$1200 epochs (of 2000 total).
    \item \textbf{Stage B (Diffusion Training):} Enable diffusion and unfreeze the denoiser to learn the reverse SDE, while keeping spectral adversarial perturbation active for robustness. The remaining epochs are used for Stage B.
\end{itemize}

\subsection{Gene-Gene Correlation Analysis}
Preserving gene-gene correlations is crucial for regulatory network inference. We compute correlation matrices over highly variable genes in real and generated data and quantify alignment via GRN Spearman and normalized matrix difference. As reported in the main text, LapDDPM achieves GRN Spearman 0.61 (Spatial) and 0.59 (Multiome), indicating recovery of higher-order gene dependencies beyond marginal distributions, while outperforming SOTA baselines.

\subsection{Additional Downstream Utility}
To complement distributional metrics, we also evaluate whether LapDDPM supports downstream biological analyses on PBMC3K. For rare-cell augmentation, we downsample pDCs to 50 cells (1.5\%) and train a classifier with synthetic augmentation; LapDDPM improves held-out F1 to 0.79, compared with 0.51 without augmentation, 0.63 for scVI, and 0.67 for scDiffusion. For differential expression consistency between CD4+ and CD8+ T-cells, LapDDPM recovers 91\% of the top-200 real DE genes, compared with 74\% for scVI and 81\% for scDiffusion. These results further support that LapDDPM preserves biologically actionable structure rather than only marginal similarity.

\iffalse

% Original CHIL sample guidelines retained for reference.
\maketitle

\begin{abstract}
This is the abstract for this article. If you are making your code
available, \emph{do not link to it in the abstract since many indexing
services will automatically remove or redact the link}. Instead,
we are requiring every paper to have an initial statement on data and
code availability right after the abstract.
\end{abstract}

\paragraph*{Data and Code Availability}
This initial paragraph is \textbf{mandatory}. Briefly state what data you
use (including citations if appropriate) and whether the data are
available to other researchers.\footnote{An example data availability
statement: This paper uses the MIMIC-III dataset
\citep{johnson2016mimic}, which is available on the PhysioNet repository
\citep{johnson2016physionet}.}
If you are not sharing code, you must explicitly state that you are not
making your code available. If you are making your code available, then
at the time of submission for review, please include your code as
supplemental material or as a code repository link; in either case, your
code must be anonymized. If your paper is accepted, then you should
de-anonymize your code for the camera-ready version of the paper. \emph{If
you do not include this data and code availability statement for your
paper, or you provide code that is not anonymized at the time of
submission, then your paper will be desk-rejected.} Your experiments later
could refer to this initial data and code availability statement if it is
helpful (e.g., to avoid restating what data you use).

\paragraph*{Institutional Review Board (IRB)}
This initial paragraph is \textbf{mandatory}. If your research requires IRB
approval or has been designated by your IRB as Not Human Subject
Research, then for the camera-ready version of the paper, you must
provide IRB information (and at the time of submission for review, you
can say that this IRB information will be provided if the paper is
accepted). If your research does not require IRB approval, then you
must state this to be the case. 

\section{Introduction}
\label{sec:intro}

This is a sample article that uses the \textsf{jmlr} class.  Please
follow the guidelines in this sample document as it can help to reduce
complications when combining the articles into a book. Please avoid
using obsolete commands, such as \verb|\rm|, and obsolete packages,
such as \textsf{epsfig}.\footnote{See
\url{http://www.ctan.org/pkg/l2tabu}} Some packages that are known
to cause problems for the production editing process are checked for
by the \textsf{jmlr} class and will generate an error. (If you want
to know more about the production editing process, have a look at
the video tutorials for the production editors at
\url{http://www.dickimaw-books.com/software/makejmlrbookgui/videos/}.)

Please also ensure that your document will compile with PDF\LaTeX.
If you have an error message that's puzzling you, first check for it
at the UK TUG FAQ
\url{https://texfaq.org/FAQ-man-latex}.  If
that doesn't help, create a minimal working example (see
\url{https://www.dickimaw-books.com/latex/minexample/}) and post
to somewhere like \TeX\ on StackExchange
(\url{http://tex.stackexchange.com/}) or the \LaTeX\ Community Forum
(\url{http://www.latex-community.org/forum/}).

\begin{note}
This is an numbered theorem-like environment that was defined in
this document's preamble.
\end{note}

\subsection{Sub-sections}

Sub-sections are produced using \verb|\subsection|.

\subsubsection{Sub-sub-sections}

Sub-sub-sections are produced using \verb|\subsubsection|.

\paragraph{Sub-sub-sub-sections}

Sub-sub-sub-sections are produced using \verb|\paragraph|.
These are unnumbered with a running head.

\subparagraph{Sub-sub-sub-sub-sections}

Sub-sub-sub-sub-sections are produced using \verb|\subparagraph|.
These are unnumbered with a running head.

\section{Cross-Referencing}

Always use \verb|\label| and \verb|\ref| (or one of the commands
described below) when cross-referencing.  For example, the next
section is Section~\ref{sec:math} but you can also refer to it using
\sectionref{sec:math}. The \textsf{jmlr} class
provides some convenient cross-referencing commands:
\verb|\sectionref|, \verb|\equationref|, \verb|\tableref|,
\verb|\figureref|, \verb|\algorithmref|, \verb|\theoremref|,
\verb|\lemmaref|, \verb|\remarkref|, \verb|\corollaryref|,
\verb|\definitionref|, \verb|\conjectureref|, \verb|\axiomref|,
\verb|\exampleref| and \verb|\appendixref|. The argument of these
commands may either be a single label or a comma-separated list
of labels. Examples:

Referencing sections: \sectionref{sec:math} or
\sectionref{sec:intro,sec:math} or
\sectionref{sec:intro,sec:math,sec:tables,sec:figures}.

Referencing equations: \equationref{eq:trigrule} or
\equationref{eq:trigrule,eq:df} or
\equationref{eq:trigrule,eq:f,eq:df,eq:y}.

Referencing tables: \tableref{tab:operatornames} or
\tableref{tab:operatornames,tab:example} or
\tableref{tab:operatornames,tab:example,tab:example-booktabs}.

Referencing figures: \figureref{fig:nodes} or
\figureref{fig:nodes,fig:teximage} or
\figureref{fig:nodes,fig:teximage,fig:subfigex} or
\figureref{fig:circle,fig:square}.

Referencing algorithms: \algorithmref{alg:gauss} or
\algorithmref{alg:gauss,alg:moore} or
\algorithmref{alg:gauss,alg:moore,alg:net}.

Referencing theorem-like environments: \theoremref{thm:eigenpow},
\lemmaref{lem:sample}, \remarkref{rem:sample}, 
\corollaryref{cor:sample}, \definitionref{def:sample},
\conjectureref{con:sample}, \axiomref{ax:sample} and
\exampleref{ex:sample}.

Referencing appendices: \appendixref{apd:first} or
\appendixref{apd:first,apd:second}.

\section{Equations}
\label{sec:math}

The \textsf{jmlr} class loads the \textsf{amsmath} package, so
you can use any of the commands and environments defined there.
(See the \textsf{amsmath} documentation for further
details.\footnote{Either \texttt{texdoc amsmath} or
\url{http://www.ctan.org/pkg/amsmath}})

Unnumbered single-lined equations should be displayed using
\verb|\[| and \verb|\]|. For example:
\[E = m c^2\]
or you can use the \texttt{displaymath} environment:
\begin{displaymath}
E = m c^2
\end{displaymath}
Numbered single-line equations should be displayed using the
\texttt{equation} environment. For example:
\begin{equation}\label{eq:trigrule}
\cos^2\theta + \sin^2\theta \equiv 1
\end{equation}
This can be referenced using \verb|\label| and \verb|\equationref|.
For example, \equationref{eq:trigrule}.

Multi-lined numbered equations should be displayed using the
\texttt{align} environment.\footnote{For reasons why you 
shouldn't use the obsolete \texttt{eqnarray} environment, see
Lars Madsen, \emph{Avoid eqnarray!} TUGboat 33(1):21--25, 2012.} For example:
\begin{align}
f(x) &= x^2 + x\label{eq:f}\\
f'(x) &= 2x + 1\label{eq:df}
\end{align}
Unnumbered multi-lined equations can be displayed using the
\texttt{align*} environment. For example:
\begin{align*}
f(x) &= (x+1)(x-1)\\
&= x^2 - 1
\end{align*}
If you want to mix numbered with unnumbered lines use the
\texttt{align} environment and suppress unwanted line numbers with
\verb|\nonumber|. For example:
\begin{align}
y &= x^2 + 3x - 2x + 1\nonumber\\
&= x^2 + x + 1\label{eq:y}
\end{align}
An equation that is too long to fit on a single line can be
displayed using the \texttt{split} environment. 
Text can be embedded in an equation using \verb|\text| or
\verb|\intertext| (as used in \theoremref{thm:eigenpow}).
See the \textsf{amsmath} documentation for further details.

\subsection{Operator Names}
\label{sec:op}

Predefined operator names are listed in \tableref{tab:operatornames}.
For additional operators, either use \verb|\operatorname|,
for example $\operatorname{var}(X)$ or declare it with
\verb|\DeclareMathOperator|, for example
\begin{verbatim}
\DeclareMathOperator{\var}{var}
\end{verbatim}
and then use this new command. If you want limits that go above and
below the operator (like \verb|\sum|) use the starred versions
(\verb|\operatorname*| or \verb|\DeclareMathOperator*|).

\begin{table*}[htbp]
\floatconts
  {tab:operatornames}%
  {\caption{Predefined Operator Names (taken from 
   \textsf{amsmath} documentation)}}%
  {%
\begin{tabular}{rlrlrlrl}
\cs{arccos} & $\arccos$ &  \cs{deg} & $\deg$ &  \cs{lg} & $\lg$ &  \cs{projlim} & $\projlim$ \\
\cs{arcsin} & $\arcsin$ &  \cs{det} & $\det$ &  \cs{lim} & $\lim$ &  \cs{sec} & $\sec$ \\
\cs{arctan} & $\arctan$ &  \cs{dim} & $\dim$ &  \cs{liminf} & $\liminf$ &  \cs{sin} & $\sin$ \\
\cs{arg} & $\arg$ &  \cs{exp} & $\exp$ &  \cs{limsup} & $\limsup$ &  \cs{sinh} & $\sinh$ \\
\cs{cos} & $\cos$ &  \cs{gcd} & $\gcd$ &  \cs{ln} & $\ln$ &  \cs{sup} & $\sup$ \\
\cs{cosh} & $\cosh$ &  \cs{hom} & $\hom$ &  \cs{log} & $\log$ &  \cs{tan} & $\tan$ \\
\cs{cot} & $\cot$ &  \cs{inf} & $\inf$ &  \cs{max} & $\max$ &  \cs{tanh} & $\tanh$ \\
\cs{coth} & $\coth$ &  \cs{injlim} & $\injlim$ &  \cs{min} & $\min$ \\
\cs{csc} & $\csc$ &  \cs{ker} & $\ker$ &  \cs{Pr} & $\Pr$
\end{tabular}\par
\begin{tabular}{rlrl}
\cs{varlimsup} & $\varlimsup$ 
& \cs{varinjlim} & $\varinjlim$\\
\cs{varliminf} & $\varliminf$ 
& \cs{varprojlim} & $\varprojlim$
\end{tabular}
}
\end{table*}

\section{Vectors and Sets}
\label{sec:vec}

Vectors should be typeset using \cs{vec}. For example $\vec{x}$.
(The original version of \cs{vec} can also be accessed using
\cs{orgvec}, for example $\orgvec{x}$.)
The \textsf{jmlr} class also provides \cs{set} to typeset a
set. For example $\set{S}$.

\section{Floats}
\label{sec:floats}

Floats, such as figures, tables and algorithms, are moving
objects and are supposed to float to the nearest convenient
location. Please don't force them to go in a particular place. In
general it's best to use the \texttt{htbp} specifier and don't
put the figure or table in the middle of a paragraph (that is
make sure there's a paragraph break above and below the float).
Floats are supposed to have a little extra space above and below
them to make them stand out from the rest of the text. This extra
spacing is put in automatically and shouldn't need modifying.

If your article will later be reprinted in the Challenges for
Machine Learning, please be aware that the CiML books use a
different paper size, so if you want to resize any images use a
scale relative to the line width (\verb|\linewidth|), text width
(\verb|\textwidth|) or text height (\verb|\textheight|).

To ensure consistency, please \emph{don't} try changing the format of the caption by doing
something like:
\begin{verbatim}
\caption{\textit{A Sample Caption.}}
\end{verbatim}
or
\begin{verbatim}
\caption{\em A Sample Caption.}
\end{verbatim}
You can, of course, change the font for individual words or 
phrases, for example:
\begin{verbatim}
\caption{A Sample Caption With Some
\emph{Emphasized Words}.}
\end{verbatim}

\subsection{Tables}
\label{sec:tables}

Tables should go in the \texttt{table} environment. Within this
environment use \verb|\floatconts| (defined by \textsf{jmlr})
to set the caption correctly and center the table contents.
The location of the caption depends on the \verb|tablecaption|
setting in the document class options.

\begin{table}[htbp]
% The first argument is the label.
% The caption goes in the second argument, and the table contents
% go in the third argument.
\floatconts
  {tab:example}%
  {\caption{An Example Table}}%
  {\begin{tabular}{ll}
  \bfseries Dataset & \bfseries Result\\
  Data1 & 0.12345\\
  Data2 & 0.67890\\
  Data3 & 0.54321\\
  Data4 & 0.09876
  \end{tabular}}
\end{table}

If you want horizontal rules you can use the \textsf{booktabs}
package which provides the commands \verb|\toprule|, 
\verb|\midrule| and \verb|\bottomrule|. For example, see
\tableref{tab:example-booktabs}.

\begin{table}[hbtp]
\floatconts
  {tab:example-booktabs}
  {\caption{A Table With Horizontal Lines}}
  {\begin{tabular}{ll}
  \toprule
  \bfseries Dataset & \bfseries Result\\
  \midrule
  Data1 & 0.12345\\
  Data2 & 0.67890\\
  Data3 & 0.54321\\
  Data4 & 0.09876\\
  \bottomrule
  \end{tabular}}
\end{table}

If you really want vertical lines as well, you can't use the
\textsf{booktabs} commands as there'll be some unwanted gaps.
Instead you can use \LaTeX's \verb|\hline|, but the rows may
appear a bit cramped.  You can add extra space above or below a
row using \verb|\abovestrut| and \verb|\belowstrut|. For example,
see \tableref{tab:example-hline}. However, you might want to read
the \textsf{booktabs} documentation regarding the use of vertical
lines.

\begin{table}[htbp]
\floatconts
  {tab:example-hline}
  {\caption{A Table With Horizontal and Vertical Lines}}%
  {%
    \begin{tabular}{|l|l|}
    \hline
    \abovestrut{2.2ex}\bfseries Dataset & \bfseries Result\\\hline
    \abovestrut{2.2ex}Data1 & 0.12345\\
    Data2 & 0.67890\\
    Data3 & 0.54321\\
    \belowstrut{0.2ex}Data4 & 0.09876\\\hline
    \end{tabular}
  }
\end{table}

If you want to align numbers on their decimal point, you can
use the \textsf{siunitx} package. For example, see
\tableref{tab:example-siunitx}. For further details see the
\textsf{siunitx} documentation\footnote{Either \texttt{texdoc
siunitx} or \url{http://www.ctan.org/pkg/siunitx}}.

\begin{table}[htbp]
\floatconts
  {tab:example-siunitx}
  {\caption{A Table With Numbers Aligned on the Decimal Point}}
  {\begin{tabular}{lS[table-format=3.5]}
  \bfseries Dataset & {\bfseries Result}\\
  Data1 & 0.12345\\
  Data2 & 10.6789\\
  Data3 & 50.543\\
  Data4 & 200.09876
  \end{tabular}}
\end{table}

If the table is too wide, you can adjust the inter-column
spacing by changing the value of \verb|\tabcolsep|. For
example:
\begin{verbatim}
\setlength{\tabcolsep}{3pt}
\end{verbatim}
If the table is very wide but not very long, you can use the
\texttt{sidewaystable} environment defined in the
\textsf{rotating} package (so use \verb|\usepackage{rotating}|).
If the table is too long to fit on a page, you can use the
\texttt{longtable} environment defined in the \textsf{longtable}
package (so use \verb|\usepackage{longtable}|).

\subsection{Figures}
\label{sec:figures}

Figures should go in the \texttt{figure} environment. Within this
environment, use \verb|\floatconts| to correctly position the
caption and center the image. Use \verb|\includegraphics|
for external graphics files but omit the file extension. Do not
use \verb|\epsfig| or \verb|\psfig|. If you want to scale the
image, it's better to use a fraction of the line width rather
than an explicit length. For example, see \figureref{fig:nodes}.

\begin{figure}[htbp]
 % Caption and label go in the first argument and the figure contents
 % go in the second argument
\floatconts
  {fig:nodes}
  {\caption{Example Image}}
  {\includegraphics[width=0.5\linewidth]{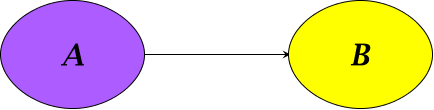}}
\end{figure}

If your image is made up of \LaTeX\ code (for example, commands
provided by the \textsf{pgf} package) you can include it using
\cs{includeteximage} (defined by the \textsf{jmlr} class). This
can be scaled and rotated in the same way as \cs{includegraphics}.
For example, see \figureref{fig:teximage}.

\begin{figure}[htbp]
\floatconts
  {fig:teximage}
  {\caption{Image Created Using \LaTeX\ Code}}
  {\includeteximage[angle=45]{images/teximage}}
\end{figure}

If the figure is too wide to fit on the page, you can use the
\texttt{sidewaysfigure} environment defined in the
\textsf{rotating} package.

Don't use \verb|\graphicspath|.\footnote{This is specific to the
\textsf{jmlr} class, not a general recommendation. The main file
that generates the proceedings or the CiML book is typically in a
different directory to the imported articles, so it modifies the
graphics path when it imports an article.} If the images 
are contained in a subdirectory, specify this when you include the image, for
example \verb|\includegraphics{figures/mypic}|.

\subsubsection{Sub-Figures}
\label{sec:subfigures}

Sub-figures can be created using \verb|\subfigure|, which is
defined by the \textsf{jmlr} class. The optional argument allows
you to provide a subcaption. The label should be placed in the
mandatory argument of \verb|\subfigure|. You can reference the
entire figure, for example \figureref{fig:subfigex}, or you can
reference part of the figure using \verb|\figureref|, for example
\figureref{fig:circle}. Alternatively you can reference the
subfigure using \verb|\subfigref|, for example
\subfigref{fig:circle,fig:square} in \figureref{fig:subfigex}.

\begin{figure}[htbp]
\floatconts
  {fig:subfigex}
  {\caption{An Example With Sub-Figures.}}
  {%
    \subfigure[A Circle]{\label{fig:circle}%
      \includegraphics[width=0.2\linewidth]{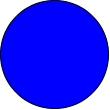}}%
    \qquad
    \subfigure[A Square]{\label{fig:square}%
      \includegraphics[width=0.2\linewidth]{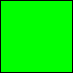}}
  }
\end{figure}

By default, the sub-figures are aligned on the baseline.
This can be changed using the second optional argument
of \verb|\subfigure|. This may be \texttt{t} (top), \texttt{c}
(centered) or \texttt{b} (bottom). For example, the subfigures
\subfigref{fig:circle2,fig:square2} in \figureref{fig:subfigex2}
both have \verb|[c]| as the second optional argument.

\begin{figure}[htbp]
\floatconts
  {fig:subfigex2}
  {\caption{Another Example With Sub-Figures.}}
  {%
    \subfigure[A Small Circle][c]{\label{fig:circle2}%
      \includegraphics[width=0.1\linewidth]{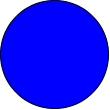}}%
    \qquad
    \subfigure[A Square][c]{\label{fig:square2}%
      \includegraphics[width=0.2\linewidth]{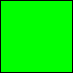}}
  }
\end{figure}

\subsection{Sub-Tables}
\label{sec:subtables}
There is an analogous command \verb|\subtable| for sub-tables.
It has the same syntax as \verb|\subfigure| described above.
You can reference the table using \verb|\tableref|, for example
\tableref{tab:subtabex} or you can reference part of the table,
for example \tableref{tab:ab}. Alternatively you can reference the
subtable using \verb|\subtabref|, for example
\subtabref{tab:ab,tab:cd} in \tableref{tab:subtabex}.

\begin{table}[htbp]
\floatconts
 {tab:subtabex}
 {\caption{An Example With Sub-Tables}}
 {%
   \subtable{%
     \label{tab:ab}%
     \begin{tabular}{cc}
     \bfseries A & \bfseries B\\
     1 & 2
     \end{tabular}
   }\qquad
   \subtable{%
     \label{tab:cd}%
     \begin{tabular}{cc}
     \bfseries C & \bfseries D\\
     3 & 4\\
     5 & 6
     \end{tabular}
   }
 }
\end{table}

By default, the sub-tables are aligned on the top.
This can be changed using the second optional argument
of \verb|\subtable|. This may be \texttt{t} (top), \texttt{c}
(centered) or \texttt{b} (bottom). For example, the sub-tables
\subtabref{tab:ab2,tab:cd2} in \tableref{tab:subtabex2}
both have \verb|[c]| as the second optional argument.

\begin{table}[htbp]
\floatconts
 {tab:subtabex2}
 {\caption{Another Example With Sub-Tables}}
 {%
   \subtable[][c]{%
     \label{tab:ab2}%
     \begin{tabular}{cc}
     \bfseries A & \bfseries B\\
     1 & 2
     \end{tabular}
   }\qquad
   \subtable[][c]{%
     \label{tab:cd2}%
     \begin{tabular}{cc}
     \bfseries C & \bfseries D\\
     3 & 4\\
     5 & 6
     \end{tabular}
   }
 }
\end{table}

\subsection{Algorithms}
\label{sec:algorithms}

Enumerated textual algorithms can be displayed using the
\texttt{algorithm} environment. Within this environment, use
\verb|\caption| to set the caption and you can use an
\texttt{enumerate} or nested \texttt{enumerate} environments.
For example, see \algorithmref{alg:gauss}. Note that algorithms
float like figures and tables.

\begin{algorithm}[htbp]
\floatconts
  {alg:gauss}%
  {\caption{The Gauss-Seidel Algorithm}}%
{%
\begin{enumerate}
  \item For $k=1$ to maximum number of iterations
    \begin{enumerate}
      \item For $i=1$ to $n$
        \begin{enumerate}
        \item $x_i^{(k)} = 
          \frac{b_i - \sum_{j=1}^{i-1}a_{ij}x_j^{(k)}
          - \sum_{j=i+1}^{n}a_{ij}x_j^{(k-1)}}{a_{ii}}$
        \item If $\|\vec{x}^{(k)}-\vec{x}^{(k-1)} < \epsilon\|$,
          where $\epsilon$ is a specified stopping criteria, stop.
      \end{enumerate}
    \end{enumerate}
\end{enumerate}
}%
\end{algorithm}

If you'd rather have the same numbering throughout the algorithm
but still want the convenient indentation of nested 
\texttt{enumerate} environments, you can use the
\texttt{enumerate*} environment provided by the \textsf{jmlr}
class. For example, see \algorithmref{alg:moore}.

\begin{algorithm}
\floatconts
  {alg:moore}%
  {\caption{Moore's Shortest Path}}%
{%
Given a connected graph $G$, where the length of each edge is 1:
\begin{enumerate*}
  \item Set the label of vertex $s$ to 0
  \item Set $i=0$
  \begin{enumerate*}
    \item \label{step:locate}Locate all unlabelled vertices 
          adjacent to a vertex labelled $i$ and label them $i+1$
    \item If vertex $t$ has been labelled,
    \begin{enumerate*}
      \item[] the shortest path can be found by backtracking, and 
      the length is given by the label of $t$.
    \end{enumerate*}
    otherwise
    \begin{enumerate*}
      \item[] increment $i$ and return to step~\ref{step:locate}
    \end{enumerate*}
  \end{enumerate*}
\end{enumerate*}
}%
\end{algorithm}

Pseudo code can be displayed using the \texttt{algorithm2e}
environment. This is defined by the \textsf{algorithm2e} package
(which is automatically loaded) so check the \textsf{algorithm2e}
documentation for further details.\footnote{Either \texttt{texdoc
algorithm2e} or \url{http://www.ctan.org/pkg/algorithm2e}}
For an example, see \algorithmref{alg:net}.

\begin{algorithm2e}
\caption{Computing Net Activation}
\label{alg:net}
% older versions of algorithm2e have \dontprintsemicolon instead
% of the following:
%\DontPrintSemicolon
% older versions of algorithm2e have \linesnumbered instead of the
% following:
%\LinesNumbered
\KwIn{$x_1, \ldots, x_n, w_1, \ldots, w_n$}
\KwOut{$y$, the net activation}
$y\leftarrow 0$\;
\For{$i\leftarrow 1$ \KwTo $n$}{
  $y \leftarrow y + w_i*x_i$\;
}
\end{algorithm2e}

\section{Description Lists}

The \textsf{jmlr} class also provides a description-like 
environment called \texttt{altdescription}. This has an
argument that should be the widest label in the list. Compare:
\begin{description}
\item[add] A method that adds two variables.
\item[differentiate] A method that differentiates a function.
\end{description}
with
\begin{altdescription}{differentiate}
\item[add] A method that adds two variables.
\item[differentiate] A method that differentiates a function.
\end{altdescription}

\section{Theorems, Lemmas etc}
\label{sec:theorems}

The following theorem-like environments are predefined by
the \textsf{jmlr} class: \texttt{theorem}, \texttt{example},
\texttt{lemma}, \texttt{proposition}, \texttt{remark}, 
\texttt{corollary}, \texttt{definition}, \texttt{conjecture}
and \texttt{axiom}. You can use the \texttt{proof} environment
to display the proof if need be, as in \theoremref{thm:eigenpow}.

\begin{theorem}[Eigenvalue Powers]\label{thm:eigenpow}
If $\lambda$ is an eigenvalue of $\vec{B}$ with eigenvector
$\vec{\xi}$, then $\lambda^n$ is an eigenvalue of $\vec{B}^n$
with eigenvector $\vec{\xi}$.
\begin{proof}
Let $\lambda$ be an eigenvalue of $\vec{B}$ with eigenvector
$\xi$, then
\begin{align*}
\vec{B}\vec{\xi} &= \lambda\vec{\xi}
\intertext{premultiply by $\vec{B}$:}
\vec{B}\vec{B}\vec{\xi} &= \vec{B}\lambda\vec{\xi}\\
\Rightarrow \vec{B}^2\vec{\xi} &= \lambda\vec{B}\vec{\xi}\\
&= \lambda\lambda\vec{\xi}\qquad
\text{since }\vec{B}\vec{\xi}=\lambda\vec{\xi}\\
&= \lambda^2\vec{\xi}
\end{align*}
Therefore true for $n=2$. Now assume true for $n=k$:
\begin{align*}
\vec{B}^k\vec{\xi} &= \lambda^k\vec{\xi}
\intertext{premultiply by $\vec{B}$:}
\vec{B}\vec{B}^k\vec{\xi} &= \vec{B}\lambda^k\vec{\xi}\\
\Rightarrow \vec{B}^{k+1}\vec{\xi} &= \lambda^k\vec{B}\vec{\xi}\\
&= \lambda^k\lambda\vec{\xi}\qquad
\text{since }\vec{B}\vec{\xi}=\lambda\vec{\xi}\\
&= \lambda^{k+1}\vec{\xi}
\end{align*}
Therefore true for $n=k+1$. Therefore, by induction, true for all
$n$.
\end{proof}
\end{theorem}

\begin{lemma}[A Sample Lemma]\label{lem:sample}
This is a lemma.
\end{lemma}

\begin{remark}[A Sample Remark]\label{rem:sample}
This is a remark.
\end{remark}

\begin{corollary}[A Sample Corollary]\label{cor:sample}
This is a corollary.
\end{corollary}

\begin{definition}[A Sample Definition]\label{def:sample}
This is a definition.
\end{definition}

\begin{conjecture}[A Sample Conjecture]\label{con:sample}
This is a conjecture.
\end{conjecture}

\begin{axiom}[A Sample Axiom]\label{ax:sample}
This is an axiom.
\end{axiom}

\begin{example}[An Example]\label{ex:sample}
This is an example.
\end{example}

\section{Color vs Grayscale}
\label{sec:color}

It's helpful if authors supply grayscale versions of their
images in the event that the article is to be incorporated into
a black and white printed book. With external PDF, PNG or JPG
graphic files, you just need to supply a grayscale version of the
file. For example, if the file is called \texttt{myimage.png},
then the gray version should be \texttt{myimage-gray.png} or
\texttt{myimage-gray.pdf} or \texttt{myimage-gray.jpg}. You don't
need to modify your code. The \textsf{jmlr} class checks for
the existence of the grayscale version if it is print mode 
(provided you have used \verb|\includegraphics| and haven't
specified the file extension).

You can use \verb|\ifprint| to determine which mode you are in.
For example, in \figureref{fig:nodes}, the 
\ifprint{dark gray}{purple} ellipse represents an input and the
\ifprint{light gray}{yellow} ellipse represents an output.
Another example: {\ifprint{\bfseries}{\color{red}}important text!}

You can use the class option \texttt{gray} to see how the
document will appear in gray scale mode. \textcolor{blue}{Colored
text} will automatically be converted to gray scale in print mode.

The \textsf{jmlr} class loads the \textsf{xcolor}
package, so you can also define your own colors. For example:
\ifprint
  {\definecolor{myred}{gray}{0.5}}%
  {\definecolor{myred}{rgb}{0.5,0,0}}%
\textcolor{myred}{XYZ}.

The \textsf{xcolor} class is loaded with the \texttt{x11names}
option, so you can use any of the x11 predefined colors (listed
in the \textsf{xcolor} documentation\footnote{either 
\texttt{texdoc xcolor} or \url{http://www.ctan.org/pkg/xcolor}}).

\section{Citations and Bibliography}
\label{sec:cite}

The \textsf{jmlr} class automatically loads \textsf{natbib}
and automatically sets the bibliography style, so you don't need to
use \verb|\bibliographystyle|.
This sample file has the citations defined in the accompanying
BibTeX file \texttt{chil-sample.bib}. For a parenthetical
citation use \verb|\citep|. For example
\citep{guyon-elisseeff-03}. For a textual citation use
\verb|\citet|. For example \citet{guyon2007causalreport}. 
Both commands may take a comma-separated list, for example
\citet{guyon-elisseeff-03,guyon2007causalreport}.

These commands have optional arguments and have a starred
version. See the \textsf{natbib} documentation for further
details.\footnote{Either \texttt{texdoc natbib} or
\url{http://www.ctan.org/pkg/natbib}}

The bibliography is displayed using \verb|\bibliography|.

\acks{Acknowledgments go here \emph{but should only appear in the
camera-ready version of the paper if it is accepted}.
Acknowledgments do not count toward the paper page limit.}

\bibliography{chil-sample}

\appendix

\section{First Appendix}\label{apd:first}

This is the first appendix.

\section{Second Appendix}\label{apd:second}

This is the second appendix.
\fi

\end{document}